\documentclass{article}

\usepackage[final]{neurips_2025}

\usepackage[utf8]{inputenc} 
\usepackage[T1]{fontenc}    
\usepackage{hyperref}       
\usepackage{url}            
\usepackage{booktabs}       
\usepackage{amsfonts}       
\usepackage{nicefrac}       
\usepackage{microtype}      
\usepackage{xcolor}         

\usepackage{amsmath}
\usepackage{array}
\usepackage{bm}
\usepackage{multirow}
\usepackage{setspace}
\usepackage{subcaption}
\usepackage{caption}
\usepackage{pifont}
\usepackage{times}
\usepackage{arydshln}
\usepackage{wrapfig, lipsum, booktabs}
\usepackage{float}
\usepackage{utfsym}
\usepackage{wrapfig}
\usepackage[normalem]{ulem}

\usepackage{amssymb}  
\usepackage{xcolor}

\newcommand{\blackcheck}{\checkmark}

\usepackage{algorithm}
\usepackage{algpseudocode}

\newcommand{\hiddensection}[1]{%
  \addtocontents{toc}{\protect\setcounter{tocdepth}{-10}}%
  \section{#1}%
  \addtocontents{toc}{\protect\setcounter{tocdepth}{2}}%
}

\newcommand{\hiddensubsection}[1]{%
  \addtocontents{toc}{\protect\setcounter{tocdepth}{1}}
  \subsection{#1}%
  \addtocontents{toc}{\protect\setcounter{tocdepth}{2}}
}

\newlength{\originalintextsep}
\BeforeBeginEnvironment{wrapfigure}{%
  \setlength{\originalintextsep}{\intextsep} 
  \setlength{\intextsep}{0pt}                
}
\AfterEndEnvironment{wrapfigure}{%
  \setlength{\intextsep}{\originalintextsep} 
}

\title{Wide-Horizon Thinking and Simulation-Based Evaluation for Real-World LLM Planning \\with Multifaceted Constraints}

\author{Dongjie Yang\textsuperscript{\rm 1}, 
    Chengqiang Lu\textsuperscript{\rm 2},
    Qimeng Wang\textsuperscript{\rm 2},
    \\
\textbf{
    Xinbei Ma\textsuperscript{\rm 1},
    Yan Gao\textsuperscript{\rm 2},
    Yao Hu\textsuperscript{\rm 2},
    Hai Zhao\textsuperscript{\rm 1}}\textsuperscript{\rm ,}\thanks{\;\;Corresponding author.} \\
    \textsuperscript{\rm 1}Shanghai Jiao Tong University, 
    \textsuperscript{\rm 2} Xiaohongshu Inc.\\
    \textsuperscript{\rm 1}\texttt{\{djyang.tony@,sjtumaxb@,zhaohai@cs.\}sjtu.edu.cn},\\
    \textsuperscript{\rm 2}\texttt{\{lusuo,haoli9,yadun,xiahou\}@xiaohongshu.com}\\
}

\begin{document}

\maketitle

\begin{abstract}
Unlike reasoning, which often entails a deep sequence of deductive steps, complex real-world planning is characterized by the need to synthesize a broad spectrum of parallel and potentially conflicting information and constraints. For example, in travel planning scenarios, it requires the integration of diverse real-world information and user preferences. While LLMs show promise, existing methods with long-horizon thinking struggle with handling multifaceted constraints, leading to suboptimal solutions. Motivated by the challenges of real-world travel planning, this paper introduces the Multiple Aspects of Planning (MAoP), empowering LLMs with "wide-horizon thinking" to solve planning problems with multifaceted constraints. Instead of direct planning, MAoP leverages the strategist to conduct pre-planning from various aspects and provide the planning blueprint for planners, enabling strong inference-time scalability by scaling aspects to consider various constraints. In addition, existing benchmarks for multi-constraint planning are flawed because they assess constraints in isolation, ignoring causal dependencies within the constraints, e.g, travel planning, where past activities dictate future itinerary. To address this, we propose Travel-Sim, an agent-based benchmark assessing plans via real-world simulation, thereby inherently resolving these causal dependencies. This paper advances LLM capabilities in complex planning and offers novel insights for evaluating sophisticated scenarios through simulation.
\end{abstract}

\hiddensection{Introduction}
\label{sec:introduction}
Large Language Models (LLMs) \cite{wang2023planandsolvepromptingimprovingzeroshot,erdogan2025plan,shridhar2021alfworldaligningtextembodied,huang2024understandingplanningllmagents} have shown significant promise in planning by generating action sequences to achieve specified goals. However, transitioning from controlled environments to the complexities of real-world planning presents a formidable challenge \cite{xie2024travelplanner,chen2024travelagent,huang2024understandingplanningllmagents}, defined by the need to manage a multitude of diverse and simultaneous constraints. Prevailing methods \cite{yao2022react,prasad2024adaptasneededdecompositionplanning} often rely on task decomposition, a linear and sequential methodology that breaks a complex problem into simpler sub-tasks. This approach is effective in domains with limited constraints, such as in GUI automation \cite{nguyen2025guiagentssurvey, wang2025guiagentsfoundationmodels}, but its feasibility diminishes sharply in real-world scenarios where constraints are deeply interconnected. The failure of this step-by-step strategy highlights a fundamental mismatch between the problem-solving approach and the non-linear nature of the challenge itself.

The cognitive model underpinning this sequential approach is analogous to that of logical and mathematical reasoning \cite{shao2024deepseekmathpushinglimitsmathematical,guan2025rstarmathsmallllmsmaster,ahn2024largelanguagemodelsmathematical}. These domains epitomize long-horizon thinking: a deep, step-by-step deductive process that the entire chain of reasoning funnels toward a limited set of deterministic outcomes. Real-world planning \cite{chen2024travelagent,hao2025largelanguagemodelssolve,zhang2024askbeforeplanproactivelanguageagents}, however, demands a fundamentally different cognitive paradigm. It requires an LLM to move beyond linear deduction to instead simultaneously integrate multifaceted constraints and heterogeneous information, necessitating wide-horizon thinking from various aspects rather than long-horizon thinking. Success hinges less on the deductive soundness of a sequence and more on the holistic feasibility of a plan that concurrently satisfies a wide array of interconnected variables.

In this paper, we explore the limitations and potential of LLMs in real-world planning through the lens of the travel planning scenario. We conduct preliminary experiments to investigate the LLM's zero-shot capability of solving the travel planning problem. Alongside long-horizon baselines, we implement a simple wide-horizon approach based on aspect-aware question decomposition. Our results indicate that LLMs struggle to address real-world planning problems using long-horizon thinking, performing notably better with even a naive wide-horizon method.

Our preliminary experiments also reveal that this naive wide-horizon approach, which relies on decomposing the request into various aspects, has its own significant limitations: 1) lack of inter-aspect associations; 2) dependence on well-crafted artifactual guidance; 3) poor inference-time scalability. To address these limitations, we propose \textbf{"Multiple Aspects of Planning" (MAoP)}. In MAoP, instead of directly planning, we introduce a strategist to conduct the pre-planning using a two-stage strategy. The strategist first analyzes the context and \textbf{decomposes} the planning request into various aspects to be considered. For each aspect, the strategist provides short guidance for the planner to further analyze subsequently. In the second \textbf{routing} stage, the strategist integrates the independent aspects into a coherent planning blueprint, leading to better inference-time scaling performance with more aspects considered. In the actual planning of MAoP, the planner concentrates on guidance of a single aspect per dialogue turn by following the pre-planning blueprint, progressively constructing a comprehensive planning process via multiple turns of dialogue.

\begin{figure}[t]
    \centering
    \includegraphics[width=0.9\linewidth]{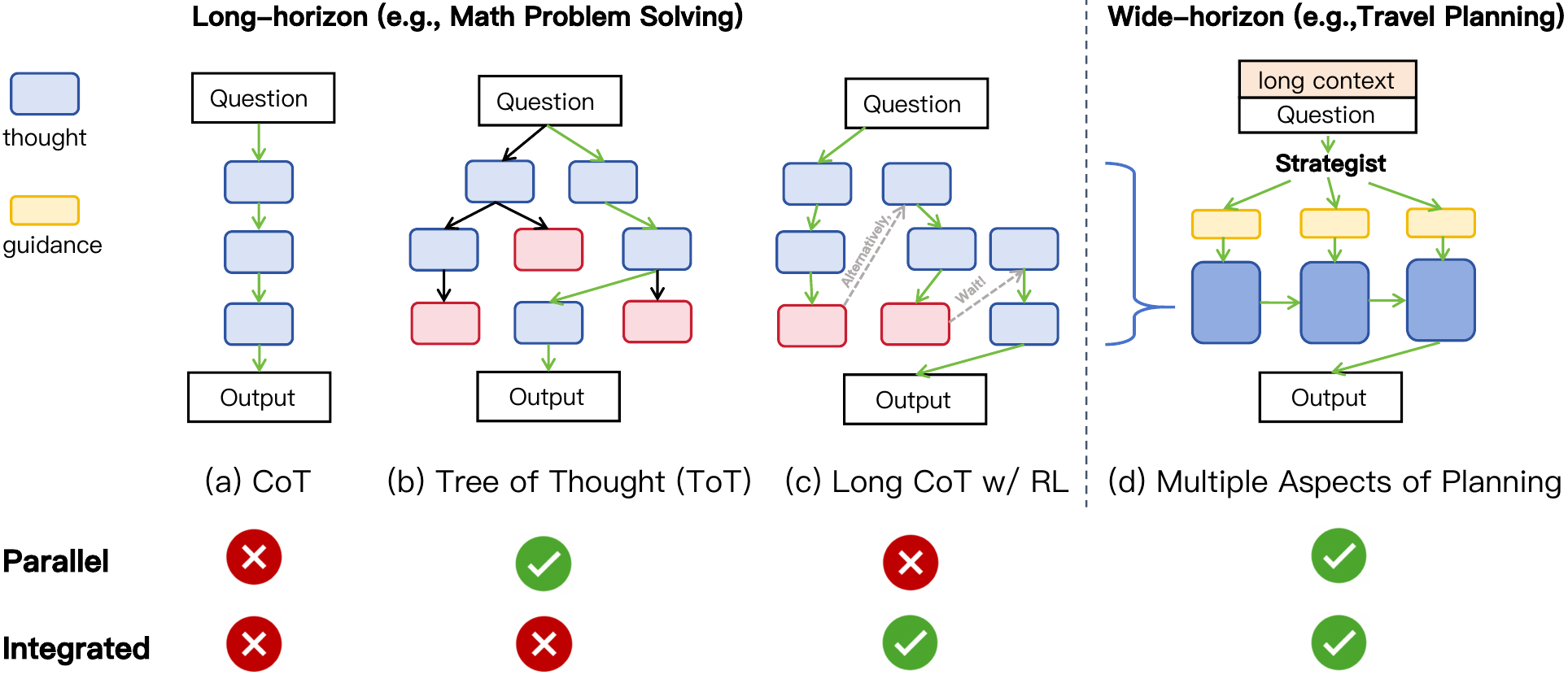}
    \caption{The comparison between long-horizon and wide-horizon thinking reveals distinct cognitive approaches. While long-horizon thinking involves deep exploration of a single reasoning trajectory, wide-horizon thinking incorporates heterogeneous information and constraints in long contexts by considering various aspects. It necessitates \textbf{parallel} consideration of multiple dimensions, which are subsequently \textbf{integrated} to generate comprehensive outputs.}
    \label{fig:comparizon_horizon}
    \vspace{-10pt}
\end{figure}

While MAoP improves real-world planning, the evaluation of multi-constraint planning remains a major hurdle. The challenges (e.g, from travel planning) are twofold: 1) The subjective nature of travel planning means there is no universal optimal solution, as users weigh constraints like cost and convenience differently, making objective evaluation elusive. 2) The causal dependency within a journey means a single failure can dynamically violate a cascade of subsequent, interconnected constraints, rendering simple, static evaluations inadequate. Previous studies \cite{xie2024travelplanner, chen2024travelagent, shao2024chinatravelrealworldbenchmarklanguage} introduce constraint pass rates as metrics, which only reflect the partial feasibility of the plan and the satisfaction of coarse-grained user requirements. These metrics ignore the real-world influence and causal consistency in real travel scenarios, poorly reflecting actual feasibility.

As the proof is in the pudding, we propose a novel agent-based evaluation framework to simulate a trip based on the real-world environment. At its core, an LLM-powered traveler agent executes a plan within an event-driven sandbox. By leveraging live traffic data from maps and qualitative insights from travel blogs, we enable the simulation to organically capture the dynamics and unforeseen events of real-world scenarios. To better emulate real-world individuals, the traveler agent is meticulously designed with diverse personas and a dynamic stamina engine that simulates physical exertion. Beyond the "static" metrics (constraint pass rates), we introduce "dynamic" metrics that the traveler evaluates the plan based on their experience, offering multi-granularity feedback to assess the experience at multiple levels.

Our contributions can be concluded as follows:
\begin{itemize}
    \item We propose MAoP to enhance wide-horizon thinking capabilities for solving real-world planning with multifaceted constraints. 
    \item 
    We propose a simulation-based evaluation framework to evaluate the feasibility and appeal of the travel plan, offering novel insights for assessing complex scenarios. 

\end{itemize}

\hiddensection{Preliminaries}
\hiddensubsection{Long-Horizon vs. Wide-Horizon}
As shown in Figure \ref{fig:comparizon_horizon},  previous studies \cite{wang2023planandsolvepromptingimprovingzeroshot, yao2023treethoughtsdeliberateproblem, yao2022react} have primarily centered on developing CoT methods and their variants to solve planning and reasoning problems requiring deep exploration along a single trajectory. Deepseek-R1 \cite{guo2025deepseekr1} further enhances the long-horizon reasoning ability by leveraging Reinforcement Learning. 
In contrast, real-world planning such as travel planning necessitates LLMs to: 1) extract pertinent information from long contexts (e.g., tour guides, spatial information, traveler information, etc); 2) conduct deliberate thinking over multifaceted constraints (e.g., real-world constraints and preferences). Travel planning problem does not require the model to reason deeply, but necessitates considering multiple aspects simultaneously in a wide-horizon view. In this paper, we investigate the potential of conducting wide-horizon thinking compared to long-horizon thinking in real-world planning.

\begin{wraptable}{r}{7cm}
\fontsize{9}{11}\selectfont
    \centering
    \caption{The comparison between long-horizon thinking and wide-horizon thinking in travel planning by evaluating the Feasibility (FEA) and Personalization (PER) scores in Travel-Sim.}
    \begin{tabular}{lcc}
    \toprule
    \textbf{Method}  & \textbf{FEA}  & \textbf{PER}  \\
    \midrule
    Qwen 2.5-32B \cite{yang2024qwen25} & & \\
    \quad w/ Artificial Guidance + CoT  & 23.3 & 36.2 \\
    \quad w/ Artificial Guidance + Plan\&Solve  & 25.0 & 39.7 \\
    \quad w/ Artificial Guidance + Wide & \textbf{31.9} & \textbf{44.1} \\
    \quad w/ Self-Gen. Guidance + Wide & 29.4 & 42.0 \\
    \midrule
    DeepSeek-R1 \cite{guo2025deepseekr1} & & \\
    \quad w/ Artificial Guidance + CoT & 52.6 &  62.5 \\
    \quad w/ Artificial Guidance + Plan\&Solve & 57.4 &  64.2 \\
    \quad w/ Artificial Guidance + Wide & \textbf{58.9} & \textbf{68.0} \\
    \quad w/ Self-Gen. Guidance + Wide & 55.2 & 64.1 \\
    \bottomrule
    \end{tabular}
    \label{tab:long_wide}
\end{wraptable}

\hiddensubsection{Wide-Horizon Thinking with Aspect-Aware Guidance}
We conducted a preliminary experiment to compare the efficacy of long-horizon versus wide-horizon thinking on real-world planning with multifaceted constraints. The input provided to the model comprises two components: a rich context and a structured guidance prompt. The context includes real-world information such as traveler profiles, travel blogs, and spatial data (see Appendix \ref{app:preprocess} for processing details). The guidance is a carefully crafted instruction that outlines the key aspects to consider when generating a travel plan. To establish our long-horizon thinking baselines, we use this guidance to prompt two methods: zero-shot CoT and the Plan-and-Solve framework \cite{wang2023planandsolvepromptingimprovingzeroshot}, which relies on task decomposition.
Our approach to naive wide-horizon thinking is centered on aspect-based decomposition. Critically, unlike prior works \cite{perez2020unsupervisedquestiondecompositionquestion,erdogan2025plan} that decompose a question into a sequence of simpler sub-tasks, our method breaks the guidance into multiple aspects designed to be considered concurrently. The LLM then independently analyzes each aspect and finally synthesizes the insights from these parallel analyses to generate an integrated output. In Table \ref{tab:long_wide}, we find that naive wide-horizon thinking with aspect-aware guidance significantly improves the travel plan quality with better feasibility and personalization scores. Although the wide-horizon thinking demonstrates significant potential, well-designed artificial guidance can further enhance the performance over self-generated guidance. More details can be checked in Section \ref{sec:experiment} and Appendix \ref{app:preliminary}.

\hiddensection{Wide-Horizon Thinking on Real-world Planning}
Based on our preliminary experiments, we find that while the naive wide-horizon thinking approach shows promise, it still exhibits several key deficiencies: 1) Although aspect-aware thinking is concurrent, its independence inherently limits the capture of inter-aspect associations; 2) the artificial guidance demonstrates superiority, yet its deliberate design remains challenging to implement; 3) as the number of aspects increases, the performance can not be scaled with more inference-time compute. Therefore, we propose the Multiple Aspects of Planning (MAoP), a method that leverages a strategist to conduct the pre-planning to address these deficiencies in naive wide-horizon thinking.

\begin{figure}
    \centering
    \includegraphics[width=1.0\linewidth]{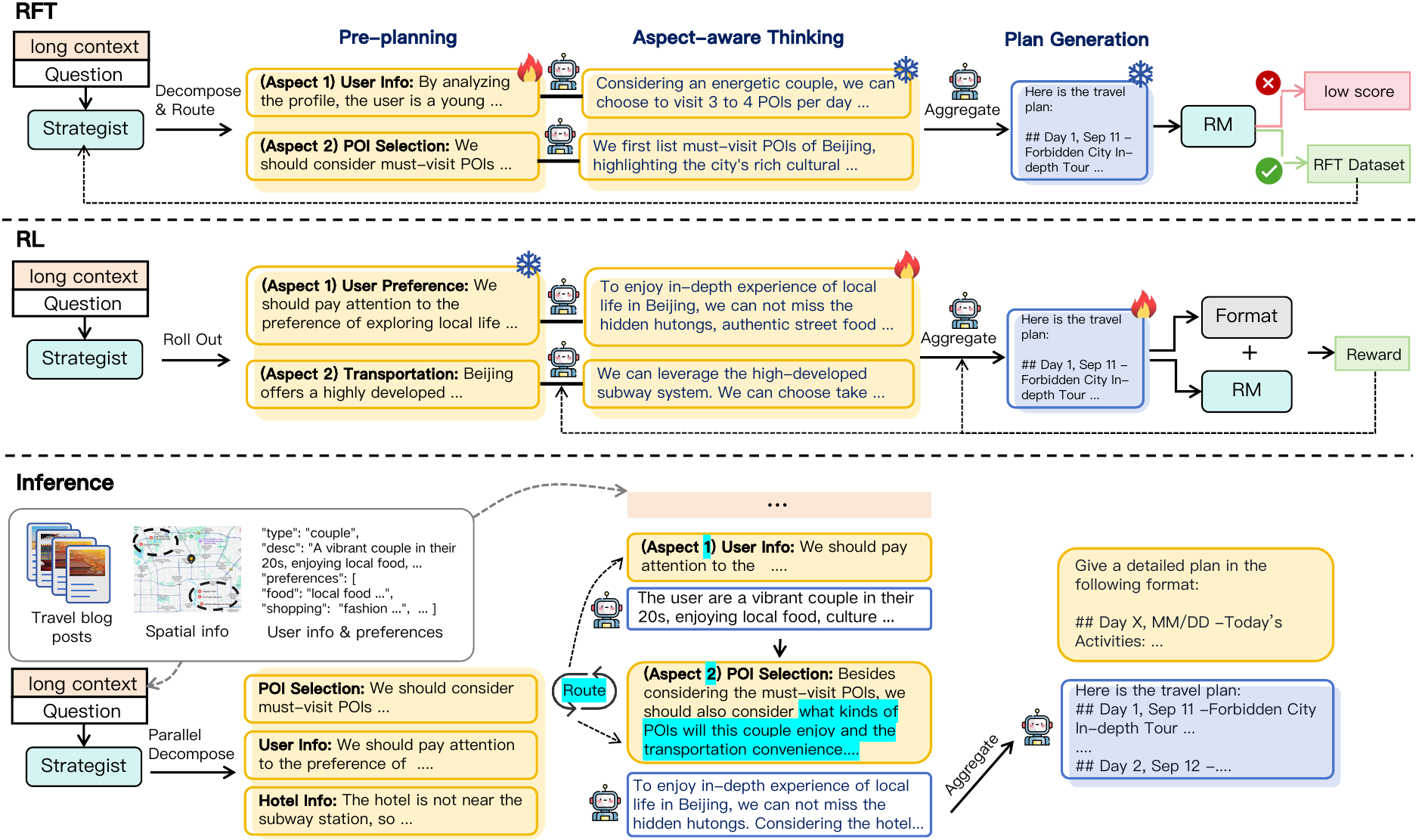}
    \caption{The overview of the MAoP training and inference process. }
    \label{fig:process}
\end{figure}

\hiddensubsection{MAoP Training}
The training of MAoP consists of three stages: reward model training, rejection sampling finetuning for the strategist \footnote{We do not further implement RL on the strategist. The RL pipeline has to go through a frozen planner to get rewards, making it hard to optimize.}, and RL training for the planner. As shown in Figure \ref{fig:process}, we mainly focus on the last two stages and provide more details in Appendix \ref{app:maop}.

\paragraph{Rejection Sampling Finetuning (RFT) for Strategist}
In MAoP, as shown in Figure \ref{fig:process}, the strategist mainly does two things in pre-planning: 1) \textbf{decomposing} the original instruction and generating the aspect-aware guidance; 2) \textbf{routing} the planning trajectories to capture inter-aspect associations. These core operations solve the mentioned deficiencies in naive wide-horizon thinking. To train such a strategist, we implement RFT while keeping the planner frozen. For each request, we prompt the strategist to first conduct pre-planning for $N$ times, and subsequently the planner to generate final plans. The rejection strategy is to reject the trajectory where all $N$ plans, as defined in Eq. \ref{eq:reward}, fall below a predefined threshold.

\paragraph{RL Training for Planner}
Following an initial cold-start RFT, we conduct RL training using GRPO \cite{shao2024deepseekmathpushinglimitsmathematical} on the planner to further improve aspect-aware thinking ability. To mitigate reward hacking, we design a multi-dimensional reward function. It is primarily guided by the PER score (a composite of five criteria, see Sec \ref{sec:eval_metric} and Appendix \ref{app:maop}) and also includes an auxiliary reward for proper plan formatting. The overall function is defined as follows:

\begin{equation}
\label{eq:reward}
R_{overall} =
\begin{cases}
  2(R_{PER} - 0.5), & \text{if the format is correct} \\
  2(R_{PER} - 0.5) - 1, & \text{if the format is incorrect}
\end{cases}
, \quad \text{where } R_{PER} \in [0, 1].
\end{equation}

\hiddensubsection{MAoP Inference}
\subsubsection{Pre-Planning}
\paragraph{Decomposition} 
As shown in Figure \ref{fig:process}, the strategist accepts the long context and question as the input and decomposes the planning request into various aspects. For each aspect, the strategist generates concise guidance to instruct the planner how to conduct a thorough analysis. By parallel sampling the strategist multiple times, we can derive a large amount of aspect-guidance pairs.

\paragraph{Routing} Different from the naive wide-horizon thinking that treats the aspects independently, the strategist additionally selects and aggregates the aspects to route the best planning blueprint. From the experiments in Appendix \ref{app:naive_scale}, we find that if we directly increase the number of aspects by sampling the strategist more times, the planner only benefits from considering 3 \textasciitilde 5 aspects without further inference-time scaling capability. It is because more aspects introduce more information but also more noise. To address this, the strategist aggregates the number of aspects into a smaller number and constructs a coherent planning blueprint. As shown in Figure \ref{fig:process}, subsequent aspect guidance is determined through the influence of preceding multiple aspects. The routing process shifts the burden of considering numerous aspects from the planner to the strategist, thus enabling the \textbf{scalability of considering more aspects to further improve the wide-horizon thinking}.

\subsubsection{Aspect-Aware Thinking}
Based on the planning blueprint constructed by the strategist, the planner sequentially conducts thinking over the aspects in a coherent multi-turn dialogue. With aspect-specific guidance, the planner can conduct a more focused and in-depth analysis over the long context from this aspect. After multiple turns of profound analysis, the planner produces the final plan based on the previous wide-horizon thoughts in the last turn of the conversation.

\hiddensubsection{MAoP Distillation for One-Step Wide-Horizon Thinking}
Implementing MAoP is complex, involving two models and a multi-turn planning process. We accelerate and simplify this process by distillation. To create high-quality training data for distillation, we employ a powerful teacher model, composed of a strategist and a planner, to generate MAoP samples. From these generated planning trajectories, we extract the strategist's aspect-specific guidance and then compress the entire aspect-aware thinking and the final aggregation into a single, consolidated output. By finetuning on this distilled data, the model learns to execute complex MAoP planning in a single inference step. This capability is what we term one-step wide-horizon thinking.

\hiddensection{Causal Evaluation via Agent-based Simulation}
As shown in Table \ref{tab:bench_comparison}, previous benchmarks typically rely on static, rule-based metrics, such as pass rates for individual constraints. However, this approach overlooks a fundamental truth: travel is a dynamic, causal process, not a static checklist. Each event, from a delayed train to physical exhaustion on the first day, directly impacts the feasibility and enjoyment of the rest of the journey. By neglecting these critical causal dependencies, existing benchmarks fail to adequately evaluate a plan's real-world viability and its capacity to meet a traveler's evolving personal needs.

We introduce Travel-Sim, a novel benchmark framework utilizing agent-based simulation. In this framework, a traveler agent, embodied by the advanced Gemini 2.5-Pro-Exp-0325 \cite{google2024gemini}, simulates a journey according to a given travel plan. Throughout the simulation, the agent provides continuous, experience-driven feedback. This dynamic, simulation-based approach inherently resolves the causal dependency issues of static benchmarks while capturing personalized, multi-granularity evaluations from an authentic traveler's perspective.

\begin{table}[ht]
\centering
\caption{Comparison of evaluation metrics for travel planning between different benchmarks.}
\label{tab:bench_comparison}
\fontsize{9}{11}\selectfont
\begin{tabular}{lcccc}
\toprule
Method  & Rule-based & LLM-Judge & Multi-Granularity & Causality \\
\midrule
TravelPlanner \cite{xie2024travelplanner} & \blackcheck & & \blackcheck & \\
UnsatChristmas \cite{hao2024largelanguagemodelssolve} & \blackcheck & & \blackcheck & \\
TravelAgent \cite{chen2024travelagent} & & \blackcheck  & & \\
ITINERA \cite{tang2024synergizing}  & \blackcheck  & \blackcheck & \blackcheck  & \\
ChinaTravel \cite{shao2024chinatravelrealworldbenchmarklanguage} & \blackcheck  & & \blackcheck & \\
\midrule
Travel-Sim \textbf{(ours)}  & \blackcheck & \blackcheck & \blackcheck & \blackcheck \\
\bottomrule
\end{tabular}
\end{table}

\hiddensubsection{Travel Experience Simulation}
To simulate a realistic travel experience, we build a sandbox to provide travelers with any real-world information they need, such as time, location, transportation, and even sightseeing experiences from blog posts. Similar to role-playing, we set up detailed profiles for traveler agents, including group sizes, character types, ages, genders, budgets, and preferences. Moreover, we design a stamina engine for the traveler. Stamina is closely related to character types (e.g., young people vs. elderly, family w/ baby vs. family w/o baby) and has a significant impact on the travel experience. The travelers of different types have their own rules for stamina exertion when encountering different events.

\paragraph{Event-Driven State Transition} The traveler follows the plan $p$ to travel in the sandbox. We define the traveler's state as $c_n=\{t, l, s, o, e\}$ at step $n$, where $t$, $l$, $s$, $o$, $e$ represent the current \emph{time}, \emph{location}, \emph{stamina}, \emph{outlay}, and the ongoing \emph{event}. The traveler employs a policy $\pi (a_n | c_1, c_2, \cdots, c_{n-1}, p)$, where action $a_n \in \mathcal{A}$ and action space $\mathcal{A}$ covers basic actions such as \emph{transiting}, \emph{resting}, \emph{dining}, and \emph{sightseeing}. Each action the traveler takes leads to a new event, which transitions the previous state to the new one. Similar to ReAct \cite{yao2022react}, the traveler agent first thinks over the current situation from the traveler's perspective and then makes a decision on the next action.

\paragraph{Real-world Information Integration}
When coming to a new city, the traveler usually starts the travel at the train station or airport. The traveler can choose an action; in most cases, the first step is to proceed to the hotel. We utilize a map API to provide various modes of transportation as references. The traveler comprehensively considers stamina (being tired from the long flight), schedule (next event in the plan), and budget to make a decision. For example, if the traveler opts to take the metro to the hotel, the environment updates the traveler's state with arrival time, new location (hotel), new stamina, etc. Among the events, the simulation of activities like sightseeing or shopping, due to the various experiences and interactions, presents a more significant challenge. Although the traveler agent can not physically visit the POI, we utilize a special event agent to generate how the traveler would do the sightseeing by referring to the real experiences in the travel blog posts.

\hiddensubsection{Multi-granularity Evaluation by Traveler}
\paragraph{Evaluation Process}
To capture detailed feedback, we implement a multi-granularity evaluation mechanism. Travelers assess their experience across five core dimensions: experience (ex), interest (it), arrangement (ar), stamina (st), and cost (co) (detailed in Appendix \ref{app:multig_eval}). These evaluations are collected at three distinct levels of granularity: per-POI (after each POI visit), per-day (at the conclusion of each day), and per-trip (upon completion of the entire journey).

\paragraph{Dataset Construction}
We construct a variety of traveler profiles, including 16 distinct types of travelers. Each differs in terms of group size, age, gender, stamina level, and preferences. We carefully select 7 Chinese cities that are ideal for tourism as destinations. We have 112 different distinct \{traveler, destination, duration\} combinations. See more details in Appendix \ref{app:data}.

\hiddensection{Experiment}
\label{sec:experiment}
\hiddensubsection{Evaluation Setup}

\paragraph{Methods}
For the baseline, we reuse the setup of preliminary experiments, including zero-shot long-horizon thinking and naive wide-horizon thinking. We add an additional baseline that we implement RL to train the models with \emph{Long/Artifact.} setting using the same training dataset as the MAoP. This additional baseline is to compare the finetuned performance between the long-horizon thinking and the MAoP method. For the MAoP, we train the Qwen 2.5-7B \cite{yang2024qwen25}, Qwen 2.5-32B \cite{yang2024qwen25}, and Deepseek-R1-Distill Qwen-7B \cite{guo2025deepseekr1} as the strategists and the planners, respectively. We also include Deepseek-R1 \cite{guo2025deepseekr1} and Gemini-2.5-Pro-Exp-03-25 \cite{google2024gemini} as additional zero-shot planners. During the MAoP inference, the strategists only route no more than 8 aspects into the planning blueprint for fair comparison.

To simplify MAoP for one-step wide-horizon thinking, we distill the MAoP synthetic data to train Qwen 2.5-3B \cite{yang2024qwen25}, Llama 3.2-3B \cite{grattafiori2024llama3}, and Deepseek-R1-Distill Qwen-7B. The teachers in the distillation are the Deepseek-R1-Distill Qwen-7B as the strategist and Gemini 2.5-Pro-Exp-0325 as the planner. More training details can be checked in Appendix \ref{app:maop}.

\begin{figure}[t]
    \centering
    \includegraphics[width=1.0\linewidth]{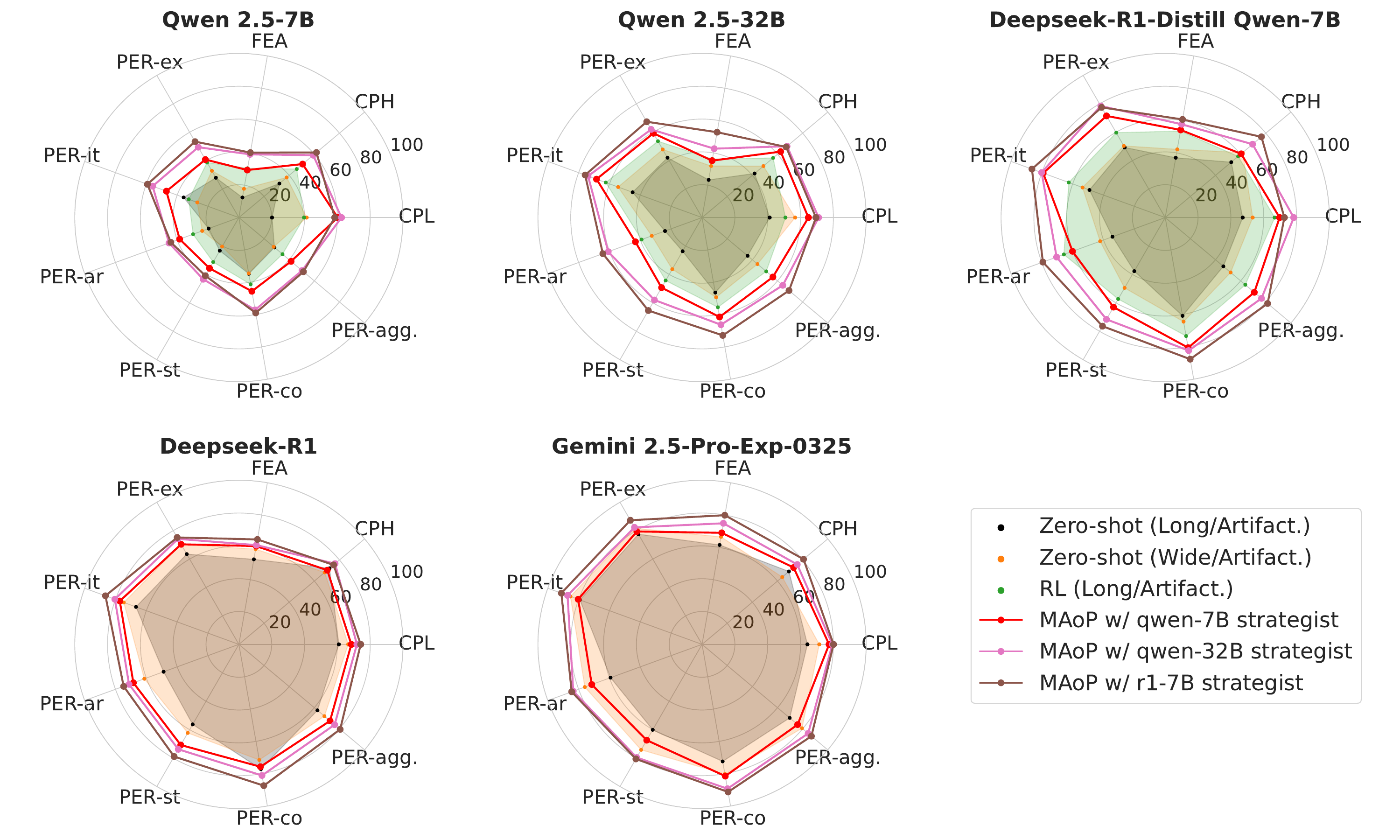}
    \caption{The comprehensive comparison results between MAoP and the baseline methods.}
    \label{fig:result}
\end{figure}

\paragraph{Metrics}
\label{sec:eval_metric}
We evaluate the MAoP on our Travel-Sim to showcase how MAoP better deals with the travel planning problem via wide-horizon thinking. We introduce four metrics (with details in Appendix \ref{app:metric}): comprehensiveness (CPH), completeness (CPL), feasibility (FEA), and personalization (PER), where the last two metrics have been used in the preliminary experiment. The first two are rule-based metrics. Comprehensiveness (CPH) evaluates how much relevant information is effectively integrated from the long context into the final plan. Completeness (CPL) evaluates if the travel plan strictly follows the formatting instructions and basic constraints according to the artifactual criteria. 

The last two metrics, feasibility (FEA) and personalization (PER), are based on the travel simulation. To evaluate the FEA, also named Travel Plan Similarity Score (TPSS), we develop an algorithm that calculates the similarity between the trajectory from the travel plan and the trajectory from the simulated travel, as shown in Figure \ref{fig:trajectory} and Algorithm \ref{alg:tpss}. This metric measures the discrepancy between what is planned and the actual execution in the simulation. If the planned trajectory is similar to the simulated one, it means the plan is more feasible. The PER is associated with the feedback of the traveler agent after the simulated travel, indicating if the plan is personalized and suitable for this traveler. The PER is a comprehensive metric, including the evaluation from multiple dimensions and granularities: 1) Multi-dimension: The traveler evaluates the travel experience from five perspectives, i.e., experience (ex), interest (it), arrangement (ar), stamina (st), and cost (co). 2) Multi-granularity: the traveler conducts evaluation from three levels (feedback after visiting every POI, finishing the whole day, and finishing the whole journey). We aggregate the scores from various dimensions and granularities of travel experience, according to Equation \ref{eq:per}, to calculate a final PER score.

\begin{table}[ht]
\centering
\caption{Comparison between MAoP distilled models and MAoP combinations.}
\label{tab:distill_comparison}
\setlength{\tabcolsep}{5pt} 
\fontsize{9}{11}\selectfont
\begin{tabular}{@{}lccccccccc@{}}
\toprule
\multirow{2}{*}{\textbf{Model}} & \multirow{2}{*}{\textbf{CPH}} & \multirow{2}{*}{\textbf{CPL}} & \multirow{2}{*}{\textbf{FEA}} & \multicolumn{6}{c}{\textbf{PER}} \\
 & & & & ex & it & ar & st & co & agg. \\
\midrule
\textbf{MAoP} &&&&&&&&& \\
\quad Qwen 2.5-7B (s.) + Qwen 2.5-32B (p.) 
& \underline{64.8} & 62.5 & 35.2 & 59.3 & \underline{68.4} & 43.2 & 49.3 & 61.5 & 56.3 \\
\quad R1-Distill 7B (s.) + R1-Distill 7B (p.) 
& 72.6 & 76.5 & 60.7 & 77.5 & 86.4 & 79.3 & \textbf{76.4} & 87.6 & 81.4 \\
\midrule
\textbf{MAoP Distillation} &&&&&&&&& \\
\quad Llama 3.2-3B (Distill) & 61.3 & 59.2 & 52.9 & 62.0 & 65.2 & 63.1 & \underline{62.5} & 70.5 & 65.7 \\
\quad Qwen 2.5-3B (Distill) & 64.2 & \underline{65.8} & \underline{53.1} & \underline{64.2} & 65.9 & \underline{63.4} & 61.4 & \underline{73.5} & \underline{66.9} \\
\quad R1-Distill Qwen-7B (Distill) & \textbf{78.2} & \textbf{79.2} & \textbf{73.7} & \textbf{84.5} & \textbf{87.2} & \textbf{83.1} & 76.0 & \textbf{90.2} & \textbf{84.2} \\
\bottomrule
\end{tabular}
\end{table}

\hiddensubsection{Result and Analysis}
\subsubsection{Benchmark Performance}
\begin{wrapfigure}{r}{6.7cm}
    \centering
    \includegraphics[width=\linewidth]{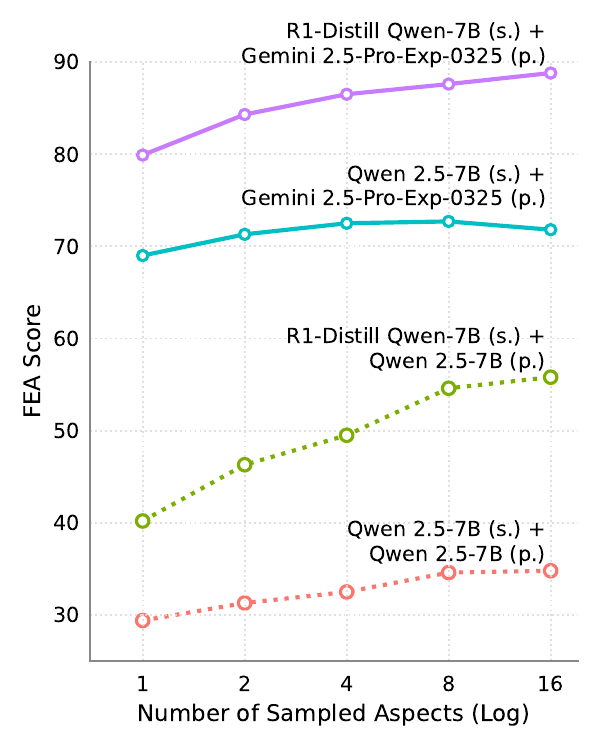} 
    \caption{We experiment two strategists (Qwen 7B \& R1-Distill Qwen-7B) and two planners (Qwen 7B \& Gemini 2.5) to showcase the scaling capability of the strategists.}
    \label{fig:scale}
\end{wrapfigure}

As illustrated in Figure \ref{fig:result}, for the finetuned models in the first row, the MAoP demonstrates a substantial performance enhancement over the long-horizon thinking (RL w/ \emph{Long/Artifact.}) baselines, achieving a remarkable 5\% to 40\% improvement across all planners. Although trained with the same dataset, MAoP achieves higher scores than RL w/ \emph{Long/Artifact.} in CPL by better constraint compliance and CPH by integrating more details from the long context to the plan. When compared with naive wide-horizon thinking (Zero-shot w/ \emph{Wide/Aritifact.}), MAoP excels especially in stronger strategists, as evidenced by higher FEA and PER scores, suggesting that the strategist plays an important role in the planning process with performance better than artifactual guidance. For advanced models as zero-shot planners, e.g., Deepseek-R1 and Gemini-2.5, the strategist also significantly boosts the performance in travel planning compared to zero-shot baselines.

\subsubsection{Inference-Time Scaling Capability of the Strategist}

As shown in Figure \ref{fig:scale}, we find that strategists can consistently improve the performance by scaling up more considered aspects. We find that even if the model size is the same, the stronger strategist with thinking (R1-Distill 7B) has better scalability, especially for advanced planners like Gemini 2.5. In contrast, the Qwen 2.5-7B model has limited scalability, because it cannot effectively route a suitable planning blueprint when dealing with increasing aspects.

\subsubsection{MAoP Distillation}
As shown in Table \ref{tab:distill_comparison}, distilling \emph{R1-Distill 7B (s.) + Gemini 2.5-Pro-Exp-0325 (p.)} enables 3B sized models to outperform the MAoP combinations even with larger model sizes. For R1-Distill 7B with the thinking mode, we also distill the thinking part of planning from the Gemini. Compared to the MAoP combination of the same models, R1-Distill 7B (Distill) achieves better performance with distilled one-step wide-horizon thinking. These results demonstrate that distillation from advanced models to smaller models achieves substantial performance improvements even without the original multi-turn MAoP process. The enhancement becomes increasingly pronounced as the capability gap between teacher and student models widens, even surpassing the MAoP training models.

\subsubsection{Emergence of Spontaneous Behaviors in Causal Travel Simulation}
\begin{figure}[H]
    \centering
    \includegraphics[width=0.9\linewidth]{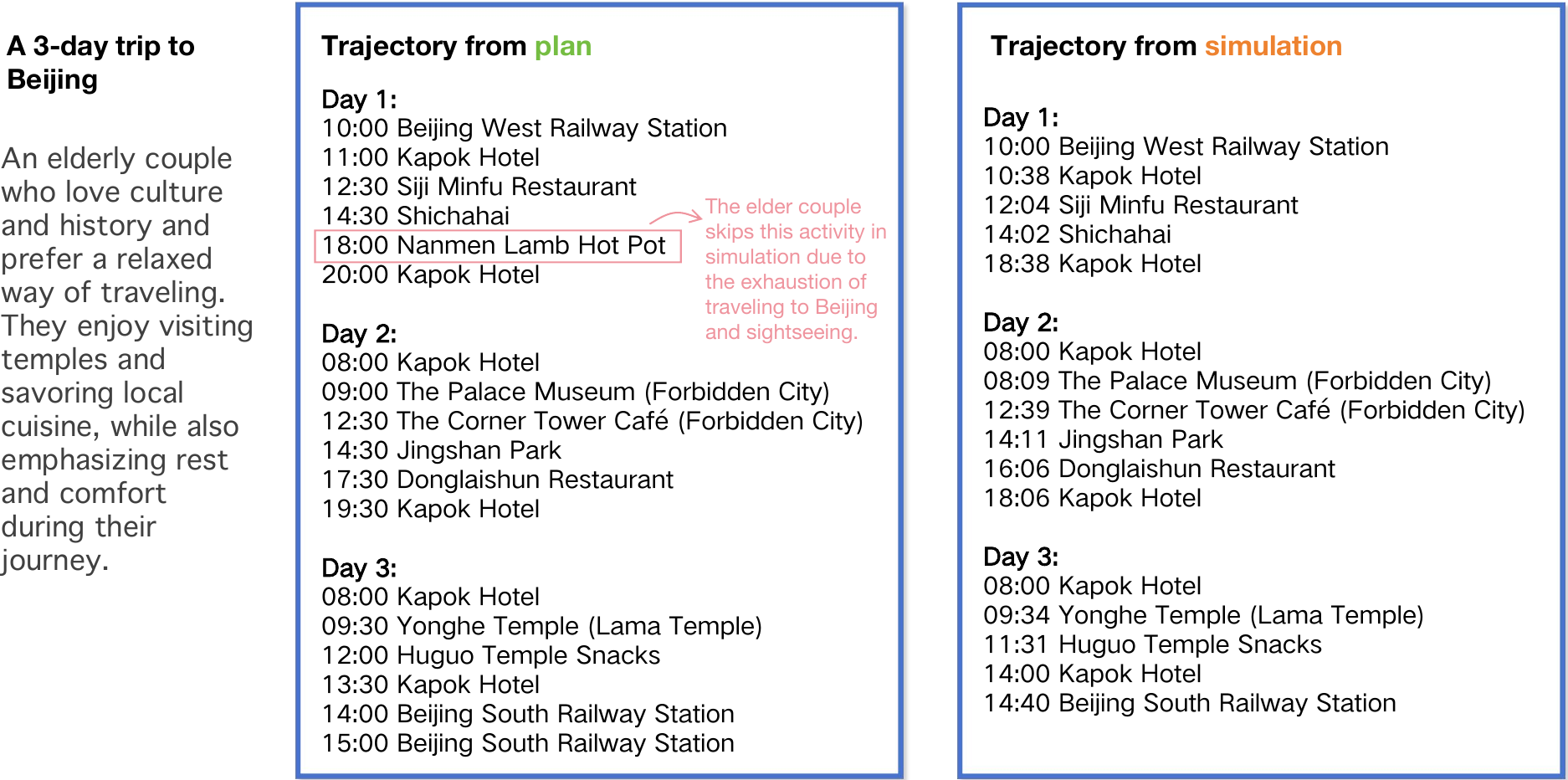}
    \caption{The simulation trajectory is not always consistent with the planned one because the traveler agent can change the subsequent itinerary based on the current situation. The \textbf{FEA} score is used to calculate the similarity of these two trajectories.}
    \label{fig:trajectory}
\end{figure}

Given the causal nature of an itinerary, the traveler agent exhibits emergent behavior by spontaneously adjusting its predetermined plan in response to unfolding events. From trajectories in Figure \ref{fig:trajectory}, this elderly couple chooses not to have dinner in Nanmen Lamb Hot Pot because they are too tired to travel to another place, especially since they just suffered from a long train trip to Beijing this morning. This indicates that neglecting causal dependencies in real-world planning evaluation can lead to significant deviations from the real-world situation.

\hiddensection{Related Work}
\paragraph{LLM Planning}
The planning capabilities of LLMs have become a key research focus due to their potential as autonomous problem-solving agents \cite{wei2023chainofthoughtpromptingelicitsreasoning, yao2022react, Wang_2024}. Recent studies have advanced these capabilities through developments in task decomposition, multi-step reasoning \cite{yao2023treethoughtsdeliberateproblem, Besta_2024}, and adaptive planning \cite{sun2023adaplanneradaptiveplanningfeedback, jutrasdubé2024adaptiveplanninggenerativemodels}. While these planning algorithms have shown promising results \cite{hao2023reasoninglanguagemodelplanning, hu2024treeplannerefficientcloselooptask}, their planning scenarios are limited to simple tasks with a single objective function. For complex real-world planning like travel planning, previous researchers focus on how to preprocess the various information and constraints to make LLM easier to understand, but few of them explore how to enhance the planning ability in this complex scenario.

\paragraph{Generative Simulation}
LLM agents begin to exhibit strong capabilities in mimicking human behaviors \cite{chuang2024demographicsaligningroleplayingllmbased, wang2023voyageropenendedembodiedagent}. Some researchers \cite{park2023generativeagentsinteractivesimulacra, wang2024userbehaviorsimulationlarge, lin2023agentsimsopensourcesandboxlarge} have been investigating the behavioral patterns of human-like LLM agents within sandbox environments, focusing on simulating human social interactions and lifestyles to study their social behaviors. Although simulation-based evaluation has become a prevalent methodology in robotics research \cite{chen2024evaluationgenerativeroboticsimulations, li2024evaluatingrealworldrobotmanipulation}, our work represents the comprehensive investigation into evaluating complex task performance using LLM agents within simulated environments.

\hiddensection{Conclusion}
We propose MAoP to enhance wide-horizon thinking for solving real-world planning problems with multifaceted constraints. In addition, we propose Travel-Sim, an evaluation benchmark that leverages agent-based simulation to offer causal and multi-granularity evaluation in travel planning scenarios. Our contributions advance the wide-horizon thinking capabilities of LLMs in real-world planning and offer novel insights for evaluating sophisticated scenarios through agent-based simulation.

\section*{Acknowledgement}
Dongjie Yang and Hai Zhao are with the Department of
Computer Science and Engineering, Shanghai Jiao Tong University; Key Laboratory of Shanghai Education Commission for Intelligent Interaction and Cognitive Engineering, Shanghai Jiao Tong University; Shanghai Key Laboratory of Trusted Data Circulation and Governance in Web3.

This paper was completed during Dongjie Yang's internship at Xiaohongshu Inc. and was supported by the Joint Research Project of Yangtze River Delta Science and Technology Innovation Community (No. 2022CSJGG1400).

\bibliography{custom}

\begin{thebibliography}{10}

\bibitem{wang2023planandsolvepromptingimprovingzeroshot}
Lei Wang, Wanyu Xu, Yihuai Lan, Zhiqiang Hu, Yunshi Lan, Roy Ka-Wei Lee, and Ee-Peng Lim.
\newblock Plan-and-solve prompting: Improving zero-shot chain-of-thought reasoning by large language models, 2023.

\bibitem{erdogan2025plan}
Lutfi~Eren Erdogan, Nicholas Lee, Sehoon Kim, Suhong Moon, Hiroki Furuta, Gopala Anumanchipalli, Kurt Keutzer, and Amir Gholami.
\newblock Plan-and-act: Improving planning of agents for long-horizon tasks.
\newblock {\em arXiv preprint arXiv:2503.09572}, 2025.

\bibitem{shridhar2021alfworldaligningtextembodied}
Mohit Shridhar, Xingdi Yuan, Marc-Alexandre Côté, Yonatan Bisk, Adam Trischler, and Matthew Hausknecht.
\newblock Alfworld: Aligning text and embodied environments for interactive learning, 2021.

\bibitem{huang2024understandingplanningllmagents}
Xu~Huang, Weiwen Liu, Xiaolong Chen, Xingmei Wang, Hao Wang, Defu Lian, Yasheng Wang, Ruiming Tang, and Enhong Chen.
\newblock Understanding the planning of llm agents: A survey, 2024.

\bibitem{xie2024travelplanner}
Jian Xie, Kai Zhang, Jiangjie Chen, Tinghui Zhu, Renze Lou, Yuandong Tian, Yanghua Xiao, and Yu~Su.
\newblock Travelplanner: A benchmark for real-world planning with language agents.
\newblock {\em arXiv preprint arXiv:2402.01622}, 2024.

\bibitem{chen2024travelagent}
Aili Chen, Xuyang Ge, Ziquan Fu, Yanghua Xiao, and Jiangjie Chen.
\newblock Travelagent: An ai assistant for personalized travel planning.
\newblock {\em arXiv preprint arXiv:2409.08069}, 2024.

\bibitem{yao2022react}
Shunyu Yao, Jeffrey Zhao, Dian Yu, Nan Du, Izhak Shafran, Karthik Narasimhan, and Yuan Cao.
\newblock React: Synergizing reasoning and acting in language models.
\newblock {\em arXiv preprint arXiv:2210.03629}, 2022.

\bibitem{prasad2024adaptasneededdecompositionplanning}
Archiki Prasad, Alexander Koller, Mareike Hartmann, Peter Clark, Ashish Sabharwal, Mohit Bansal, and Tushar Khot.
\newblock Adapt: As-needed decomposition and planning with language models, 2024.

\bibitem{nguyen2025guiagentssurvey}
Dang Nguyen, Jian Chen, Yu~Wang, Gang Wu, Namyong Park, Zhengmian Hu, Hanjia Lyu, Junda Wu, Ryan Aponte, Yu~Xia, Xintong Li, Jing Shi, Hongjie Chen, Viet~Dac Lai, Zhouhang Xie, Sungchul Kim, Ruiyi Zhang, Tong Yu, Mehrab Tanjim, Nesreen~K. Ahmed, Puneet Mathur, Seunghyun Yoon, Lina Yao, Branislav Kveton, Jihyung Kil, Thien~Huu Nguyen, Trung Bui, Tianyi Zhou, Ryan~A. Rossi, and Franck Dernoncourt.
\newblock Gui agents: A survey, 2025.

\bibitem{wang2025guiagentsfoundationmodels}
Shuai Wang, Weiwen Liu, Jingxuan Chen, Yuqi Zhou, Weinan Gan, Xingshan Zeng, Yuhan Che, Shuai Yu, Xinlong Hao, Kun Shao, Bin Wang, Chuhan Wu, Yasheng Wang, Ruiming Tang, and Jianye Hao.
\newblock Gui agents with foundation models: A comprehensive survey, 2025.

\bibitem{shao2024deepseekmathpushinglimitsmathematical}
Zhihong Shao, Peiyi Wang, Qihao Zhu, Runxin Xu, Junxiao Song, Xiao Bi, Haowei Zhang, Mingchuan Zhang, Y.~K. Li, Y.~Wu, and Daya Guo.
\newblock Deepseekmath: Pushing the limits of mathematical reasoning in open language models, 2024.

\bibitem{guan2025rstarmathsmallllmsmaster}
Xinyu Guan, Li~Lyna Zhang, Yifei Liu, Ning Shang, Youran Sun, Yi~Zhu, Fan Yang, and Mao Yang.
\newblock rstar-math: Small llms can master math reasoning with self-evolved deep thinking, 2025.

\bibitem{ahn2024largelanguagemodelsmathematical}
Janice Ahn, Rishu Verma, Renze Lou, Di~Liu, Rui Zhang, and Wenpeng Yin.
\newblock Large language models for mathematical reasoning: Progresses and challenges, 2024.

\bibitem{hao2025largelanguagemodelssolve}
Yilun Hao, Yongchao Chen, Yang Zhang, and Chuchu Fan.
\newblock Large language models can solve real-world planning rigorously with formal verification tools, 2025.

\bibitem{zhang2024askbeforeplanproactivelanguageagents}
Xuan Zhang, Yang Deng, Zifeng Ren, See-Kiong Ng, and Tat-Seng Chua.
\newblock Ask-before-plan: Proactive language agents for real-world planning, 2024.

\bibitem{shao2024chinatravelrealworldbenchmarklanguage}
Jie-Jing Shao, Xiao-Wen Yang, Bo-Wen Zhang, Baizhi Chen, Wen-Da Wei, Guohao Cai, Zhenhua Dong, Lan-Zhe Guo, and Yu~feng Li.
\newblock Chinatravel: A real-world benchmark for language agents in chinese travel planning, 2024.

\bibitem{yao2023treethoughtsdeliberateproblem}
Shunyu Yao, Dian Yu, Jeffrey Zhao, Izhak Shafran, Thomas~L. Griffiths, Yuan Cao, and Karthik Narasimhan.
\newblock Tree of thoughts: Deliberate problem solving with large language models, 2023.

\bibitem{guo2025deepseekr1}
Daya Guo, Dejian Yang, Haowei Zhang, Junxiao Song, Ruoyu Zhang, Runxin Xu, Qihao Zhu, Shirong Ma, Peiyi Wang, Xiao Bi, et~al.
\newblock Deepseek-r1: Incentivizing reasoning capability in llms via reinforcement learning.
\newblock {\em arXiv preprint arXiv:2501.12948}, 2025.

\bibitem{yang2024qwen25}
An~Yang, Baosong Yang, Beichen Zhang, Binyuan Hui, Bo~Zheng, Bowen Yu, Chengyuan Li, Dayiheng Liu, Fei Huang, Haoran Wei, et~al.
\newblock Qwen2. 5 technical report.
\newblock {\em arXiv preprint arXiv:2412.15115}, 2024.

\bibitem{perez2020unsupervisedquestiondecompositionquestion}
Ethan Perez, Patrick Lewis, Wen tau Yih, Kyunghyun Cho, and Douwe Kiela.
\newblock Unsupervised question decomposition for question answering, 2020.

\bibitem{google2024gemini}
Google.
\newblock Our most intelligent ai models, built for the agentic era.
\newblock \url{https://deepmind.google/technologies/gemini/}, 2025.

\bibitem{hao2024largelanguagemodelssolve}
Yilun Hao, Yongchao Chen, Yang Zhang, and Chuchu Fan.
\newblock Large language models can solve real-world planning rigorously with formal verification tools, 2024.

\bibitem{tang2024synergizing}
Yihong Tang, Zhaokai Wang, Ao~Qu, Yihao Yan, Kebing Hou, Dingyi Zhuang, Xiaotong Guo, Jinhua Zhao, Zhan Zhao, and Wei Ma.
\newblock Synergizing spatial optimization with large language models for open-domain urban itinerary planning.
\newblock {\em arXiv preprint arXiv:2402.07204}, 2024.

\bibitem{grattafiori2024llama3}
Aaron Grattafiori, Abhimanyu Dubey, Abhinav Jauhri, Abhinav Pandey, Abhishek Kadian, Ahmad Al-Dahle, Aiesha Letman, Akhil Mathur, Alan Schelten, Alex Vaughan, et~al.
\newblock The llama 3 herd of models.
\newblock {\em arXiv preprint arXiv:2407.21783}, 2024.

\bibitem{wei2023chainofthoughtpromptingelicitsreasoning}
Jason Wei, Xuezhi Wang, Dale Schuurmans, Maarten Bosma, Brian Ichter, Fei Xia, Ed~Chi, Quoc Le, and Denny Zhou.
\newblock Chain-of-thought prompting elicits reasoning in large language models, 2023.

\bibitem{Wang_2024}
Lei Wang, Chen Ma, Xueyang Feng, Zeyu Zhang, Hao Yang, Jingsen Zhang, Zhiyuan Chen, Jiakai Tang, Xu~Chen, Yankai Lin, Wayne~Xin Zhao, Zhewei Wei, and Jirong Wen.
\newblock A survey on large language model based autonomous agents.
\newblock {\em Frontiers of Computer Science}, 18(6), March 2024.

\bibitem{Besta_2024}
Maciej Besta, Nils Blach, Ales Kubicek, Robert Gerstenberger, Michal Podstawski, Lukas Gianinazzi, Joanna Gajda, Tomasz Lehmann, Hubert Niewiadomski, Piotr Nyczyk, and Torsten Hoefler.
\newblock Graph of thoughts: Solving elaborate problems with large language models.
\newblock {\em Proceedings of the AAAI Conference on Artificial Intelligence}, 38(16):17682–17690, March 2024.

\bibitem{sun2023adaplanneradaptiveplanningfeedback}
Haotian Sun, Yuchen Zhuang, Lingkai Kong, Bo~Dai, and Chao Zhang.
\newblock Adaplanner: Adaptive planning from feedback with language models, 2023.

\bibitem{jutrasdubé2024adaptiveplanninggenerativemodels}
Pascal Jutras-Dubé, Ruqi Zhang, and Aniket Bera.
\newblock Adaptive planning with generative models under uncertainty, 2024.

\bibitem{hao2023reasoninglanguagemodelplanning}
Shibo Hao, Yi~Gu, Haodi Ma, Joshua~Jiahua Hong, Zhen Wang, Daisy~Zhe Wang, and Zhiting Hu.
\newblock Reasoning with language model is planning with world model, 2023.

\bibitem{hu2024treeplannerefficientcloselooptask}
Mengkang Hu, Yao Mu, Xinmiao Yu, Mingyu Ding, Shiguang Wu, Wenqi Shao, Qiguang Chen, Bin Wang, Yu~Qiao, and Ping Luo.
\newblock Tree-planner: Efficient close-loop task planning with large language models, 2024.

\bibitem{chuang2024demographicsaligningroleplayingllmbased}
Yun-Shiuan Chuang, Krirk Nirunwiroj, Zach Studdiford, Agam Goyal, Vincent~V. Frigo, Sijia Yang, Dhavan Shah, Junjie Hu, and Timothy~T. Rogers.
\newblock Beyond demographics: Aligning role-playing llm-based agents using human belief networks, 2024.

\bibitem{wang2023voyageropenendedembodiedagent}
Guanzhi Wang, Yuqi Xie, Yunfan Jiang, Ajay Mandlekar, Chaowei Xiao, Yuke Zhu, Linxi Fan, and Anima Anandkumar.
\newblock Voyager: An open-ended embodied agent with large language models, 2023.

\bibitem{park2023generativeagentsinteractivesimulacra}
Joon~Sung Park, Joseph~C. O'Brien, Carrie~J. Cai, Meredith~Ringel Morris, Percy Liang, and Michael~S. Bernstein.
\newblock Generative agents: Interactive simulacra of human behavior, 2023.

\bibitem{wang2024userbehaviorsimulationlarge}
Lei Wang, Jingsen Zhang, Hao Yang, Zhiyuan Chen, Jiakai Tang, Zeyu Zhang, Xu~Chen, Yankai Lin, Ruihua Song, Wayne~Xin Zhao, Jun Xu, Zhicheng Dou, Jun Wang, and Ji-Rong Wen.
\newblock User behavior simulation with large language model based agents, 2024.

\bibitem{lin2023agentsimsopensourcesandboxlarge}
Jiaju Lin, Haoran Zhao, Aochi Zhang, Yiting Wu, Huqiuyue Ping, and Qin Chen.
\newblock Agentsims: An open-source sandbox for large language model evaluation, 2023.

\bibitem{chen2024evaluationgenerativeroboticsimulations}
Feng Chen, Botian Xu, Pu~Hua, Peiqi Duan, Yanchao Yang, Yi~Ma, and Huazhe Xu.
\newblock On the evaluation of generative robotic simulations, 2024.

\bibitem{li2024evaluatingrealworldrobotmanipulation}
Xuanlin Li, Kyle Hsu, Jiayuan Gu, Karl Pertsch, Oier Mees, Homer~Rich Walke, Chuyuan Fu, Ishikaa Lunawat, Isabel Sieh, Sean Kirmani, Sergey Levine, Jiajun Wu, Chelsea Finn, Hao Su, Quan Vuong, and Ted Xiao.
\newblock Evaluating real-world robot manipulation policies in simulation, 2024.

\bibitem{Sheng_2025}
Guangming Sheng, Chi Zhang, Zilingfeng Ye, Xibin Wu, Wang Zhang, Ru~Zhang, Yanghua Peng, Haibin Lin, and Chuan Wu.
\newblock Hybridflow: A flexible and efficient rlhf framework.
\newblock In {\em Proceedings of the Twentieth European Conference on Computer Systems}, EuroSys ’25, page 1279–1297. ACM, March 2025.

\end{thebibliography}
\bibliographystyle{unsrt}







\newpage
\appendix
\setcounter{tocdepth}{2}
\renewcommand{\contentsname}{Appendix Contents}
\tableofcontents
\newpage

\section{Limitations and Ethics Statement}
\subsection{Limitations}
\label{app:limitations}
While our proposed method (MAoP) and agent-based simulation benchmark (Travel-Sim) demonstrate significant advancements in complex travel planning and evaluation, several limitations warrant discussion.

\subsubsection{MAoP Limitations}
Though MAoP significantly enhances wide-horizon thinking with deliberate pre-planning by the strategist, MAoP still exhibits a few limitations: 1) MAoP requires the collaboration between the strategist and planner, and the conduction of aspect-aware thinking in multiple turns of dialogue. This complicated process costs more inference time than directly using CoT. To alleviate this, we propose that the MAoP distillation can compress this complicated process, gaining similar efficiency compared to CoT. 2) We find that weak strategists, especially those with sizes smaller than 7B, can cause severe performance degradation due to low-quality planning blueprint and poor inference-time scalability. The recommended way to enhance wide-horizon thinking ability for smaller models is to distill the outputs of the advanced model after MAoP training.

\subsubsection{Travel-Sim Limitations}
The Travel-Sim is the pioneer in evaluating with agent-based simulation, offering a novel solution for complex scenario evaluation, e.g., travel planning. Agent-based simulation and evaluation have several limitations that can be discussed: 1) Similar to other LLM-based evaluation benchmarks, the "traveler", which is an LLM, has the potential bias and unreproducibility over the evaluation. To mitigate this limitation, we stipulate the agent model to be Gemini-2.5-Pro-Exp-0325 \cite{google2024gemini} using greedy decoding (0 temperature) for reproducibility. We also verify the consistency between LLM-based and human-based evaluation in the Appendix \ref{app:human}, showcasing the 92\% consistency in PER score. 2) Although Travel-Sim takes causality and real-world information into account, we do not include all the emergencies that may happen in the real trip, especially some force majeure and human factors. For example, bad weather or delays caused by carelessness also significantly influence the trip. 3) Although Travel-Sim offers causal and multi-granularity evaluation, the whole evaluation process is time-consuming and costs about 20K tokens per sample. To evaluate the 112 samples of the entire dataset, we cost around \$12 by consuming about 2M tokens in total.


\subsection{Ethics Statement}
\label{app:ethics}
For the traveler profiles in the training dataset and Travel-Sim, we do not use any personal information to collect traveler profiles. The diversity of travelers is manually designed, and data is synthesized using Gemini-2.0-Pro-Exp-0205 \cite{google2024gemini}. For the sightseeing event agent that uses travel blog posts to simulate travel experience, we collect blog posts from Red Note (Xiaohongshu) that may contain personal information and copyrighted items. Therefore, people using the blog posts should respect the privacy and copyrights of the blog post owner and strictly agree to the license in Appendix \ref{app:license}.

\section{License}
\label{app:license}
By downloading or using our open-source data, you understand, acknowledge, and agree to all the terms in the following agreement.

\paragraph{ACADEMIC USE ONLY}
Any content from the Travel-Sim dataset is available for academic research purposes only. You agree not to reproduce, duplicate, copy, trade, or exploit for any commercial purposes.

\paragraph{NO DISTRIBUTION}
Respect the privacy of the personal information of the original source. Without the permission of the copyright owner, you are not allowed to perform any form of broadcasting, modification, or any other similar behavior to the data set content.

\paragraph{RESTRICTION AND LIMITATION OF LIABILITY}
In no event shall we be liable for any other damages whatsoever arising out of the use of, or inability to use this dataset and its associated software, even if we have been advised of the possibility of such damages.

\paragraph{DISCLAIMER}
You are solely responsible for legal liability arising from your improper use of the dataset content. We reserve the right to terminate your access to the dataset at any time. You should delete the Travel-Sim dataset if required.

You must comply with all terms and conditions of these original licenses, including but not limited to the Google Gemini Terms of Use, the Copyright Rules \& Policies of Red Note (Xiaohongshu). This project does not impose any additional constraints beyond those stipulated in the original licenses.

\section{Preprocessing Framework}
In this section, we introduce how to collect and preprocess the heterogeneous information as the "long context" for travel planning. \textbf{This context information serves as the information source input for both the training dataset and Travel-Sim.}

\begin{figure*}[ht]
    \centering
    \includegraphics[width=\linewidth]{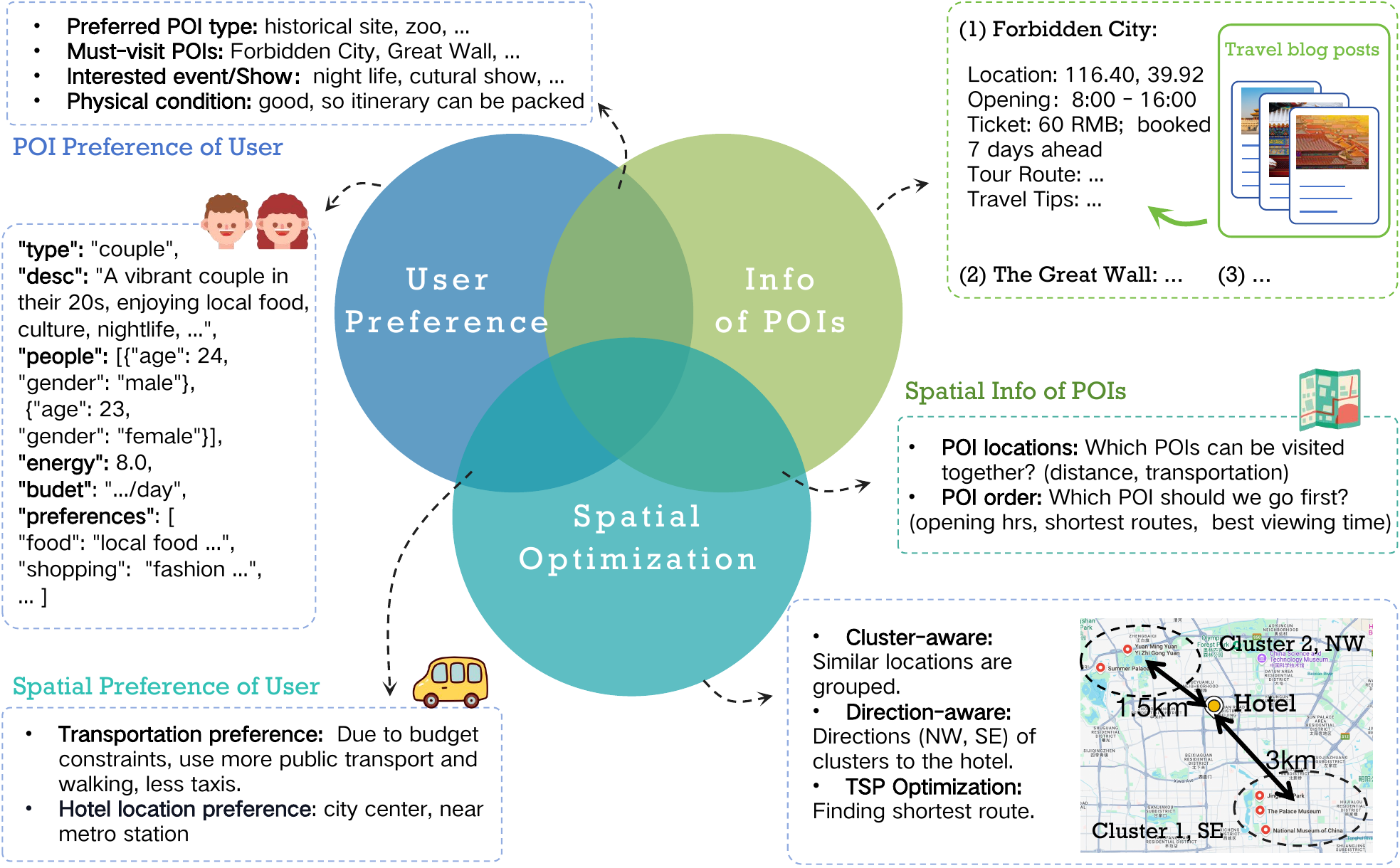}
    \caption{The decomposition of multifaceted constraints and heterogeneous information to be considered when creating a travel plan.}
    \label{fig:source}
\end{figure*}

We break the information and constraints necessary for planning into different categories shown in Figure \ref{fig:source}. We leverage our framework to collect and preprocess them for further planning. The framework consists of multiple modules that collaborate with each other.

\label{app:preprocess}
\subsection{Proactive Consultant Agent for Implicit Preferences Elicitation}
Users often provide insufficient information, making it challenging for the assistant to devise an effective plan. To address this, we not only extract preferences from user requests but also proactively ask questions to elicit implicit preferences. We design two agents to gather users' travel and accommodation preferences, respectively. The travel consultant agent collects basic trip details (group size, duration, traveler type) and integrates insights from travel blogs to identify potential challenges for travelers. The accommodation consultant first inquires about hotel preferences and then searches travel blogs and maps to identify optimal hotels. Through iterative interactions, agents can uncover implicit preferences for travel planning and accommodation.

\subsubsection{Travel Consultant Agent}
We develop an LLM-based agent to serve as a travel consultant, designed to comprehensively dig out the user's implicit preferences. We design six aspects that the travel consultant agent needs to take into account, as shown in Table \ref{tab:travel_consultant}.

\begin{table}[htbp]
\centering
\caption{Considerations for implicit preferences elicitation.}
\label{tab:travel_consultant}
\fontsize{9}{11}\selectfont
\begin{tabular}{p{3cm}p{10cm}}
\toprule
\textbf{Type} & \textbf{Description} \\
\midrule
Duration of Visit & How long does the user wish to tour, is it one day, two days, or longer? \\
\midrule
Group Composition & How many people are in the group, what is their approximate age, and are there any special groups, such as the elderly or children? \\
\midrule
Attraction Type & What type of attractions does the user hope to visit, cultural relics, natural scenery, or shopping and entertainment? \\
\midrule
Number of Attractions & Does the user want to visit as many attractions as possible, or not too many? \\
\midrule
Mode of Transportation & Under replaceable circumstances, what mode of transportation does the user prefer: walking, cycling, public transport, taxi, or driving? \\
\midrule
Special Requirements & Does the user have any special needs or preferences, such as particular requirements for shopping or sightseeing? \\
\bottomrule
\end{tabular}
\end{table}

Based on the initial user request, the travel consultant agent will proactively inquire about several questions concerning six aspects to better understand the traveler's situation. Upon gathering sufficient information across these six aspects, the agent will synthesize a comprehensive conclusion in the form of a traveler profile. The agent will subsequently present this profile to the user for review, allowing them to verify the details and provide any necessary additions or corrections through further interaction with the agent.

Instead of asking the same fixed questions, travel consultant agents have the following advantages: 1) Customized Service: we observe that the LLMs will adjust the questions based on the destination and the traveler type, and also give relevant advice to guide the user. 2) Feedback: The agent can receive feedback from the user and raise further interaction to refine the traveler profile.

\subsubsection{Hotel Consultant Agent}
The selection of the hotel is the key part of travel planning, which previous researchers seldom take into account. The hotel influences not only the accommodation experience but also the planning and logistics of the entire trip, concerning the hotel locations. We develop an LLM-based agent to serve as a hotel consultant agent to recommend hotels based on travelers' preferences.

We decompose this process into five phases: 1) \textbf{Inquiry phase}: similar to the travel consultant agent, the hotel consultant agent has to inquire about the user preferences. 2) \textbf{Online searching phase}: after understanding the hotel and POI preferences, the agent can search the internet to find the compliant hotels from the latest tour guides. 3) \textbf{Map searching phase}: the agent leverages the map API to search for hotels near the most-visited locations and conclude a list that meets the requirements. 4) \textbf{Conclusion phase}: the agent recommends the hotels from the travel blog posts and the map API. 5) \textbf{Feedback phase}: the agent receives the feedback and revises the recommendation.

\begin{table}[htbp]
\centering
\caption{Considerations for accommodation planning.}
\label{tab:accommodation_planning}
\fontsize{9}{11}\selectfont
\begin{tabular}{p{4cm}p{9cm}}
\toprule
\textbf{Type} & \textbf{Description} \\
\midrule
Hotel Consistency & Does the user prefer to stay in the same hotel throughout the trip, or change hotels according to the attractions? \\
\midrule
Accommodation Budget & What is the user's budget for accommodation? \\
\midrule
Location Preference & Does the user have any requirements for the location of the accommodation, such as being in the city center, near attractions, etc.? \\
\midrule
Facility Requirements & What facilities does the user require from the accommodation? \\
\bottomrule
\end{tabular}
\end{table}

To be specific, in the inquiry phase, we consider four aspects, as shown in Table \ref{tab:accommodation_planning}. After several proactive interactions, the agent will stop asking if it has collected enough user preferences. Besides the hotel preference, the hotel agent can leverage POI preferences and spatial information (POI clusters) from other modules. In the online searching phase, we use the search API provided by the Red Note (Xiaohongshu) app, a popular Chinese app focused on lifestyle content sharing. We use the Red Note (Xiaohongshu) search API to search for blog posts related to hotel recommendations. In the map searching phase, we use the Amap (Gaode) Map API, a Chinese map app similar to Google Maps, to search for hotels near the most-visited locations. The probable most-visited locations are predicted by the agent based on the given information. In the conclusion phase, the agent will recommend the hotels provided by the blog posts and the map API by comprehensively taking the locations and preferences into account.

\subsection{Preference-aware POI Selection}
As we have the user preferences collected by the travel consultant agent, we leverage the information to find the potential POIs and activities that will appeal to users. We first leverage LLMs to generate several search queries based on user preferences and use the Red Note (Xiaohongshu) API to search the relevant blog posts. Based on the preferences, we use LLMs to extract the POIs that match user preferences. We use the Amap (Gaode) map API to obtain the POI meta information. We finally filter and deduplicate them to get a list of candidate POIs.

\subsection{Cluster-based Spatial Optimization}
To enhance the coherence of travelers' journeys and minimize unnecessary back-and-forth travel, similar to ITINERA \cite{tang2024synergizing}, we cluster the POIs based on geographical distances from each other. This way, visitors can follow a well-planned, spatially coherent route for their tours, for example, visiting the POIs in the same cluster in one day, to enjoy a travel experience that is both efficient and pleasant. We use K-means++ to cluster the POIs, where the number of clusters is determined jointly by the number of travel days and the candidate POIs. 

After obtaining the POI clusters, we can describe the spatial distribution characteristics of POIs from two perspectives: intra-cluster features and inter-cluster relationships, to enable LLMs to understand the spatial information. 

\begin{algorithm}[h]
\caption{Find Shortest Route in the POI Cluster}
\label{alg:find_shortest_route}
\begin{algorithmic}
\State \textbf{Input:} {locations set} $L$, start location name $s$, end location name $e$
\State \textbf{Output:} shortest path $P$, shortest distance $d_{total}$, step distances $\{d_i\}_{i=1}^{n-1}$

\Function{FindShortestRoute}{$L, s, e$}
    \State Extract other locations excluding $s$ and $e$: $O \leftarrow \{loc \in L \mid loc \notin \{s, e\}\}$
    \State Let $n \leftarrow |O|$
    
    \If{$n < 6$}
        \State Compute all permutations of $O$: $\Pi \leftarrow \{\pi \mid \pi \text{ is a permutation of } O\}$
        \State Find shortest path and distance from $\Pi$:\\ $(P, d_{total}) \leftarrow \arg\min_{\pi \in \Pi} \sum_{i=1}^{|\pi|} \text{CalculateDistance}(C[\pi[i]], C[\pi[i+1]])$
    \Else
        \State Get initial path by finding nearest neighbors: $P \leftarrow \text{NearestNeighbor}(L, s, e)$
        \State Optimize path using 2-opt search: $P \leftarrow \text{TwoOptSearch}(P, C, s, e)$
        \State Calculate shortest distance: $d_{total} \leftarrow \sum_{i=1}^{|P|-1} \text{CalculateDistance}(C[P[i]], C[P[i+1]])$
    \EndIf
    
    \State Calculate step distances: $\{d_i\}_{i=1}^{|P|-1} \leftarrow \{\text{CalculateDistance}(C[P[i]], C[P[i+1]])\}_{i=1}^{|P|-1}$
    \State \textbf{return} $(P, d_{total}, \{d_i\}_{i=1}^{|P|-1})$
\EndFunction

\end{algorithmic}
\end{algorithm}

\subsubsection{Intra-cluster: Finding the Shortest Routes within the Cluster}
Since the POIs in the same cluster are relatively close to each other, travelers have the potential to visit them in sequence within a single day. In order to find the shortest route that does not retrace any part of the path, this problem can be formulated as a variant of the Traveling Salesman Problem (TSP). As the classic Traveling Salesman Problem (TSP) is an NP-hard problem in combinatorial optimization, we use brute force to solve clusters with a small number of POIs (fewer than 6), and for those with a larger number, we use the 2-opt algorithm to find an approximate solution. As shown in Algorithm \ref{alg:find_shortest_route}, we have the start location $s$ and end location $e$, which are generally set as the hotel. Locations besides $s$ and $e$ are the POIs in the same clusters. We calculate the shortest route that starts at the hotel and ends at the hotel after visiting the POIs.

\subsubsection{Inter-cluster: Relative Cluster Locations to the Hotel}
The locations of the POI clusters relative to hotels are also key information for LLM to know the spatial distribution of all POIs. We mainly calculate the direction and distance from the cluster center to the hotel, as shown in Algorithm \ref{alg:calculate_directions_distances}. To make directions easier to understand for LLMs, we can simplify directions into eight basic compass points.

\begin{algorithm}[h]
\caption{Calculate Directions and Distances}
\label{alg:calculate_directions_distances}
\begin{algorithmic}
\State \textbf{Input:} Start point $S = (\lambda_s, \phi_s)$, set of target points $\mathcal{T} = \{(T_i, C_i)\}_{i=1}^{n}$ where $C_i = (\lambda_i, \phi_i)$
\State \textbf{Output:} Set of tuples $\mathcal{O}$ containing direction $D_i$ and distance $d_i$ for each target point $T_i$

\Function{CalculateDirectionsAndDistances}{$S, \mathcal{T}$}
    \State Initialize empty set $\mathcal{O}$
    \ForAll{$(T_i, C_i) \in \mathcal{T}$}
        \State Convert coordinates to radians: $\lambda'_s, \phi'_s, \lambda'_i, \phi'_i$
        \State Calculate geodesic distance $d_i$ between $S$ and $C_i$
        \State Compute bearing $\theta'$:
            $$
            \theta' = \arctan2\left(\sin(\lambda'_i - \lambda'_s) \cdot \cos(\phi'_i), 
            \cos(\phi'_s) \cdot \sin(\phi'_i) - \sin(\phi'_s) \cdot \cos(\phi'_i) \cdot \cos(\lambda'_i - \lambda'_s)\right)
            $$
        \State Normalize bearing $\theta$: $\theta = (\theta' \cdot \frac{180}{\pi} + 360) \mod 360$
        \State Map bearing to cardinal direction $D_i$: $D_i = \text{directions}[\text{round}(\theta / 45) \mod 8]$
        \State Add tuple $(D_i, d_i)$ to $\mathcal{O}$
    \EndFor
    \State \textbf{return} $\mathcal{O}$
\EndFunction
\end{algorithmic}
\end{algorithm}

\subsection{Entire Preprocessing Workflow}
\begin{figure}[h]
    \centering
    \includegraphics[width=0.8\linewidth]{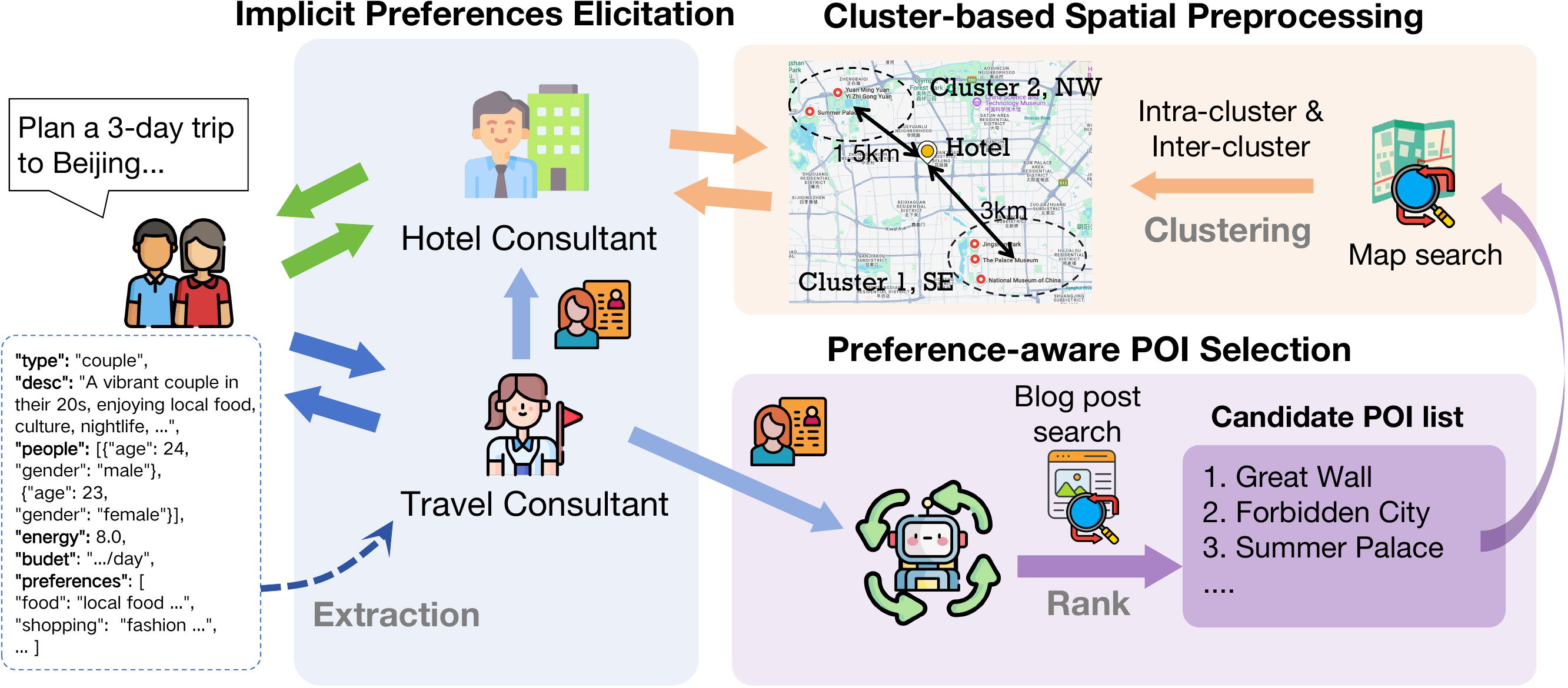}
    \caption{The entire workflow of the preprocessing framework.}
    \label{fig:framework}
\end{figure}

\paragraph{Workflow}
The preprocessing framework consists of these modules above that work together in collaboration rather than operating independently. As shown in Figure \ref{fig:framework}, we present the entire workflow of the framework. First, as the user raises a request for planning a trip, the travel consultant agent gathers user preferences and information by interacting with the user. Based on the preferences, we search the relevant blog posts and extract the candidate POIs. We perform clustering on the candidate POIs. In the next step, the hotel consultant inquires about the hotel preference and comprehensively considers all the information (user profile, hotel preference, and spatial information) to recommend hotels. After we have the location of the hotels, we conduct intra-cluster and inter-cluster analysis. 

\begin{table}[ht]
\centering
\caption{The composition of the long context input.}
\label{tab:context}
\fontsize{9}{11}\selectfont
\begin{tabular}{p{4cm}p{3cm}p{5.5cm}}
\toprule
\textbf{Type} & \textbf{Source} & \textbf{Description} \\
\midrule
User Profile \& Preferences & Travel Consultant & The information of the traveler/traveler group and corresponding preferences. \\
\midrule
Hotel Information & Hotel Consultant & The location and other information about where the user lives. \\
\midrule
POI Blog Posts & Blog Post Search & The relevant blog posts that provide sightseeing tips for each POI. \\
\midrule
Inter-Cluster Information & Spatial Preprocessing & The shortest routes within the POI cluster. \\
\midrule
Inter-Cluster Information & Spatial Preprocessing & The relative direction and distance from cluster centers to hotels. \\
\bottomrule
\end{tabular}
\end{table}

\paragraph{Final Input Context for LLMs}
After we collect and preprocess the information, we organize it as the context part of the LLM Input. To give an intuitive view, we structure the context in Table \ref{tab:context}. It is noted that for every candidate POI, we search for the most relevant blog post that provides the latest sightseeing tips. We can not depend on the intrinsic knowledge of LLMs about the POIs, which is outdated and inaccurate most of the time. The latest blog posts provide most of the information and tips for visiting the POI, for example, the optimal sightseeing routes, which help LLMs to optimize the itinerary and personalize the plan.

\section{Preliminary Experiment Details}
\label{app:preliminary}
\subsection{Detailed Setup}
\paragraph{Naive Wide-Horizon Thinking with Artifactual Guidance}
\label{app:aspect_design}
In the preliminary experiment of exploring naive wide-horizon thinking, we craft the artificial guidance that includes a series of aspects to be considered. We manually design five aspects as shown below:

\includegraphics[width=1.0\linewidth]{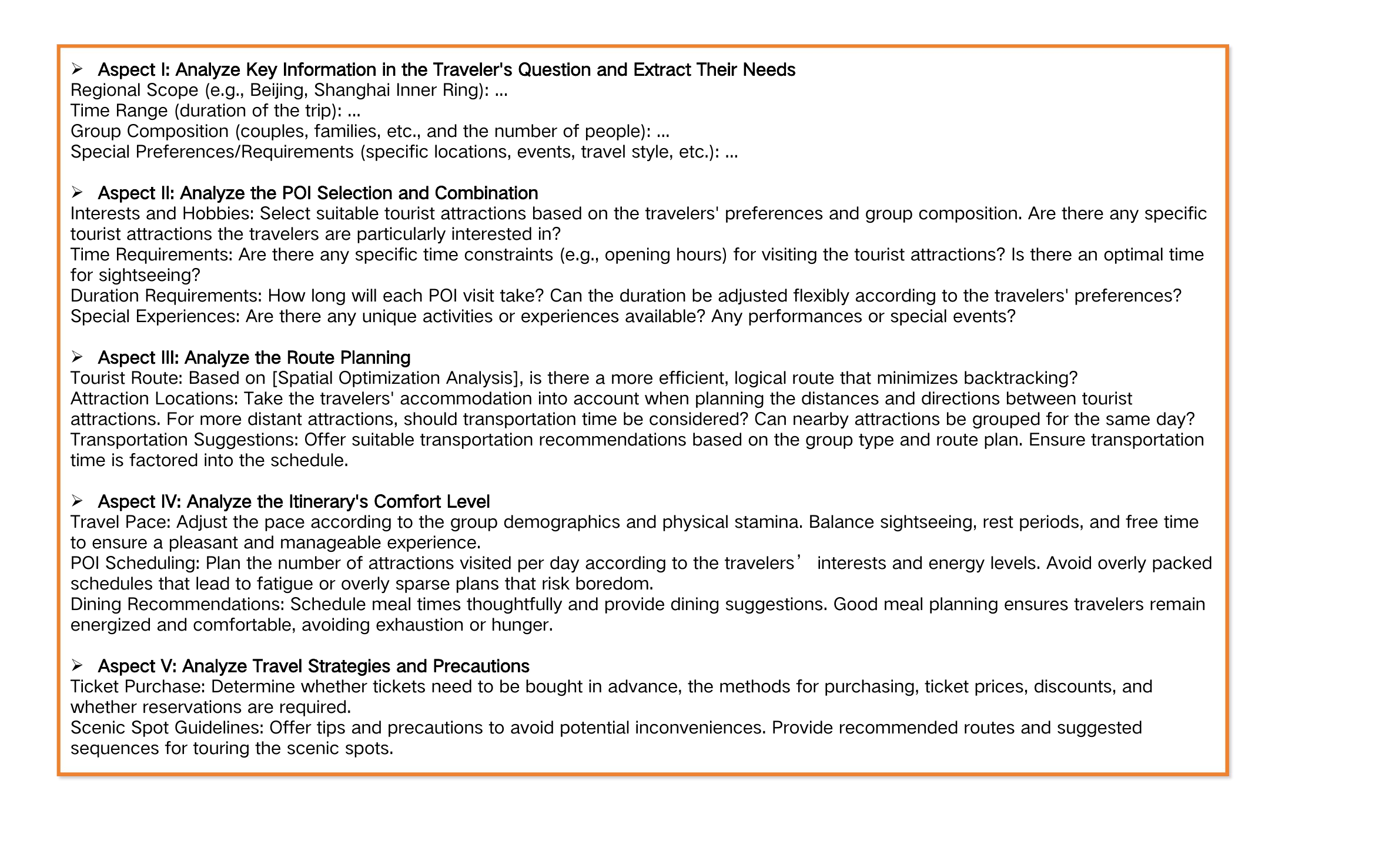}

\paragraph{Naive Wide-Horizon Thinking with Self-Generated Guidance}
Instead of using the artifactual guidance, we leverage the LLM to first analyze the aspects to be considered and generate the corresponding guidance. This process can be perceived as the zero-shot strategist, as shown below:

\includegraphics[width=1.0\linewidth]{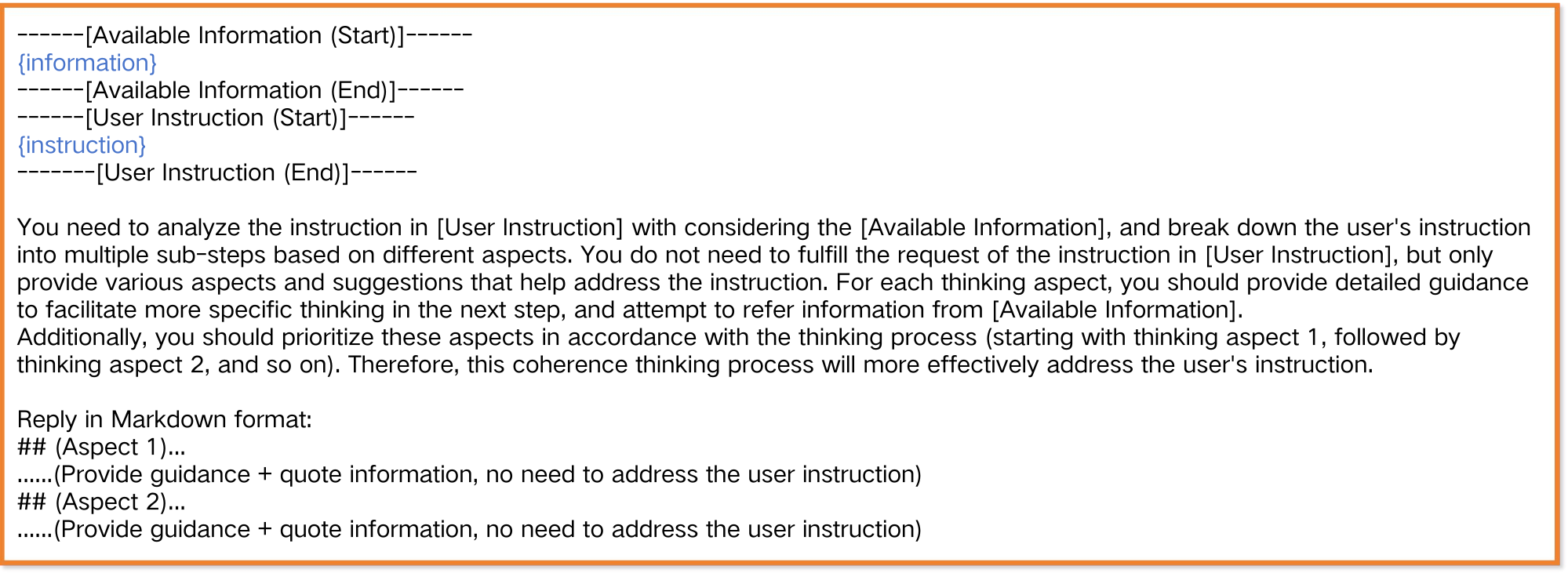}

\subsection{Scaling Capability of Naive Wide-horizon Thinking}
\label{app:naive_scale}
We investigate the potential benefit of scaling inference-time computation with more aspects for naive wide-horizon thinking. 

\begin{figure*}[ht]
    \centering
    \includegraphics[width=\linewidth]{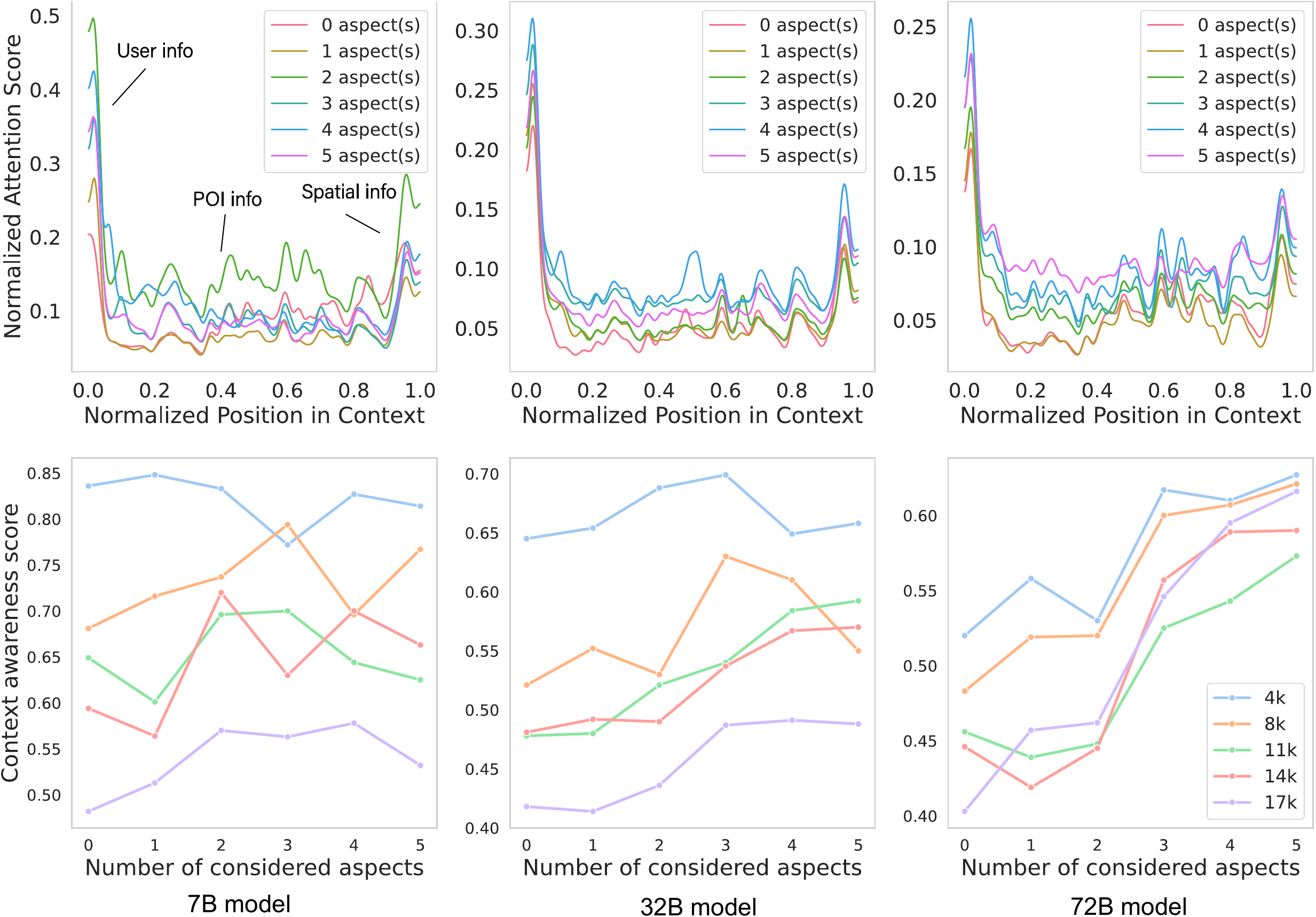}
    \caption{When LLMs consider different numbers of aspects, we analyze the attention pattern of the context (excluding the instruction) on different sizes of Qwen-2.5, varying from 7B, 32B, and 72B (from left to right). \textbf{(Upper row)} We present the distribution of attention scores across the context, aggregated from the output tokens. \textbf{(Lower row)} Additionally, we investigate the impact of context length (number of context tokens) by examining the context awareness scores. Our analysis reveals that larger models exhibit an enhanced capability for wide-horizon thinking, enabling them to focus on a broader range of information within longer contexts by considering multiple aspects simultaneously.}
    \label{fig:attn_dist}
\end{figure*}

\begin{table*}[ht]
\centering
\setlength{\tabcolsep}{5pt} 
\fontsize{9}{11}\selectfont
\begin{tabular}{ccccc|cccc|cccc}
\toprule
\multirow{2}{*}{Aspect} & \multicolumn{4}{c}{Qwen 2.5-7B} & \multicolumn{4}{c}{Qwen 2.5-32B} & \multicolumn{4}{c}{Qwen 2.5-72B}  \\
\cmidrule(lr){2-5} \cmidrule(lr){6-9} \cmidrule(lr){10-13}
& CPH $\uparrow$  & CPL $\uparrow$ & FEA $\uparrow$ & PER $\uparrow$ & CPH  & CPL  & FEA  & PER  & CPH  & CPL  & FEA  & PER    \\
\midrule
0  & 18.2  & \textbf{40.2}  & 15.2  & 27.8  
   & 39.6  & 51.1  & 21.7  & 34.6 
   & 50.4  & 62.5  & 37.0  & 44.0   \\
1  & 20.3  & 39.5  & 14.4  & 28.5  
   & 42.8  & \textbf{53.3}  & 25.4  & 33.3  
   & 51.4  & 60.3  & 37.2 & 43.4    \\
2  & 25.3  & 36.2  & \textbf{19.0}  & 30.6   
   & 47.0  & 50.1  & 31.4 & 38.7  
   & 54.9  & 60.2  & 39.2 & 46.0    \\
3  & \textbf{31.5}  & 37.3  & 16.3  & \textbf{32.6}  
  & 49.9  & 50.4 & \textbf{33.0}  & 42.3  
  & 56.5 & \textbf{62.9} & 41.5 & 49.7    \\
4 & 29.8  & 37.8  & 14.5  & 29.4  
  & \textbf{52.9}  & 47.2  & 32.0  & \textbf{43.5} 
  & 60.0  & 61.7 & 41.8 & 52.5 \\
5 & 28.3 & 38.0 & 17.8  & 27.7  
  & 52.7 & 48.8  & 31.9  & 43.1  
  & \textbf{60.3} & 59.5  & \textbf{42.1} & \textbf{53.5} \\
\bottomrule
\end{tabular}
\caption{Comparing the effectiveness of wide-horizon thinking when considering a different number of aspects by quantitative metrics.}
\label{tab:performance}
\end{table*}

\paragraph{Experiment Setup} We investigate scenarios ranging from 0 aspect (i.e., zero-shot CoT baseline without artifactual guidance) to a maximum of 5 artifactual aspects considered. We evaluate how effectively context is considered in the thinking process by analyzing attention patterns with respect to context. Besides attention patterns, we evaluate if wide-horizon thinking is effective and enhances the quality of travel plans by leveraging four metrics: CPL, CPH, FEA, and PER scores. We conduct the experiments on Qwen-2.5 \cite{yang2024qwen25} with different sizes, including 7B, 32B, and 72B.

\paragraph{Result Analysis}
In the upper part of Figure \ref{fig:attn_dist}, we show curves of the attention score distributions over the context (excluding the instruction). The attention scores are aggregated from the output tokens and then averaged across layers and attention heads. We observe that larger LLMs (34B and 72B) exhibit higher attention scores across the context when considering more aspects, indicating a strong ability to focus on relevant details. In contrast, the smaller LLM (7B) does not follow this pattern, achieving its highest scores when two aspects are considered. In addition, we introduce another indicator, dubbed context awareness score $\mathcal{S}_c$, to indicate the proportion of the most attended tokens in the context out of the total, as calculated in Equation \ref{eq:cas}: 
\begin{equation}
\label{eq:cas}
\begin{gathered}
\mathcal{S}_c = \frac{1}{n} \sum_{i=0}^{n}\mathbb{I}_{\mathcal{A}^c \in \mathrm{Top}k}(\mathcal{A}^c_{i}),
\end{gathered}   
\end{equation}
where $A^c$ denotes the attention scores over the context from the output tokens. In the lower part of Figure \ref{fig:attn_dist}, for larger models like 72B, as the context length increases, the $\mathcal{S}_c$ curves of longer length rise more steeply. It indicates that by considering more aspects, LLMs can capture finer details in longer contexts. This phenomenon emerges only when models are large enough, as smaller LLMs do not exhibit explicit improvements in $\mathcal{S}_c$ with long contexts.

As shown in Table \ref{tab:performance}, larger LLMs demonstrate enhanced travel planning capabilities as they consider more aspects, achieving progressively higher FEA and PER scores. However, this improved performance comes at the cost of formatting compliance, as evidenced by decreasing CPL scores.

\section{MAoP Experiment Details}
\label{app:maop}
\subsection{Training Details}
\subsubsection{Reward Model}
\paragraph{Training Dataset} As shown in Table \ref{tab:city}, we collect 14 diverse cities as destinations and 37 traveler types. Based on these destinations and travelers, we first synthesize 1K travel planning requests. We then separately use Gemini-2.0-Pro-Exp-0205 and Qwen 2.5-32B to directly generate travel plans. For about 2K generated plans, we conduct the simulation-based evaluation to get the PER scores as the reward score labels. 

\paragraph{Model Structure}
To avoid reward hacking, instead of directly predicting the reward (PER-agg. score), the reward model is required to generate the corresponding scores for the five criteria in the PER score and aggregate them into the final reward.

\paragraph{Training Setup} We conduct the pointwise generative reward modeling. We use the travel plan as the input. Instead of directly using the aggregated scalar score calculated in Equation \ref{eq:per}, we extract the assessment text of the traveler and the corresponding PER score after the whole journey as the output. We use 8 H800 GPUs to finetune the Qwen 2.5-7B model for 5 epochs with a learning rate of 2e-5, global batch size of 64.

\subsubsection{Strategist}
\paragraph{Training Dataset} To train a strategist with RFT, we construct the training dataset consisting of 50K samples with 120 distinct traveler types and 20 Chinese cities. We first conduct RFT on Deepseek-R1-Distill Qwen-7B to get the initial RFT data. We reject the sample that gets the average PER score lower than 40 after sampling $N=3$ times. We use 32 H800 GPUs to train Deepseek-Distill Qwen-7B for 3 epochs with a learning rate of 2e-5, global batch size of 128.

\paragraph{Training Setup} After training the Deepseek-R1-Distill Qwen-7B, we finally collect 22K RFT data with rejection sampling outputs. We reuse this 22K data to finetune the Qwen 2.5 7B and 32B. We use 8 H800 GPUs to train Qwen 2.5 7B for 3 epochs with a learning rate of 2e-5, global batch size of 64. We use 32 H800 GPUs to train the Qwen 2.5-32B strategist for 3 epochs with a learning rate of 1e-5, global batch size of 128.

\subsubsection{Planner}
\paragraph{Training Dataset} We first train the planner using the 22K data above as the cold start for subsequent RL training. To train the planner with RL, we additionally construct the training dataset consisting of 20K samples.

\paragraph{Training Setup} We conduct the standard GRPO \cite{shao2024deepseekmathpushinglimitsmathematical} using the veRL \cite{Sheng_2025} framework. We use 32 H800 GPUs to train Deepseek-R1-Distill Qwen-7B, Qwen 2.5-32B, and Qwen 2.5-7B, for 200 steps with learning rates of 1e-6, 5e-7, 1e-6, respectively, and the KL loss coefficient of 0.001. We use the KL penalty to prevent severe biases. For loss calculation, we mask the aspect-aware guidance from the strategists, preventing the planners from generating the guidance by themselves.

\subsubsection{Distillation}
\paragraph{Training Dataset} We reuse the input of the 22K data and use \emph{Deepseek-R1-Distill Qwen-7B (s.) + Gemini-2.5-Pro-Exp-0325 (p.)} to generate the output. For each sample, we sample the output 3 times. We use the reward model to filter out the output whose reward score is lower than 80. We use the filtered dataset consisting of 15K samples to train the Qwen 2.5-3B, Llama 3.2-3B, and Deepseek-R1-Distill Qwen-7B. For the first two smaller models, we discard the thinking part. For Deepseek-R1-Distill Qwen-7B, we use the thinking part from Gemini-2.5-Pro-Exp-0325. 

\paragraph{Training Setup} We use 32 H800 GPUs to train Deepseek-Distill Qwen-7B for 5 epochs with a learning rate of 2e-5, global batch size of 128. We use 8 H800 GPUs to train Qwen 2.5 3B and Llama 3.2 3B for 5 epochs with a learning rate of 2e-5, global batch size of 64.

\subsection{Inference Details}
\subsubsection{Inference for Strategist}
\paragraph{Decomposition}
The first stage of pre-planning is to decompose the planning request into multiple aspects to be considered and provide guidance for further analysis. Here is the prompt used in the RFT and inference:

\includegraphics[width=1.0\linewidth]{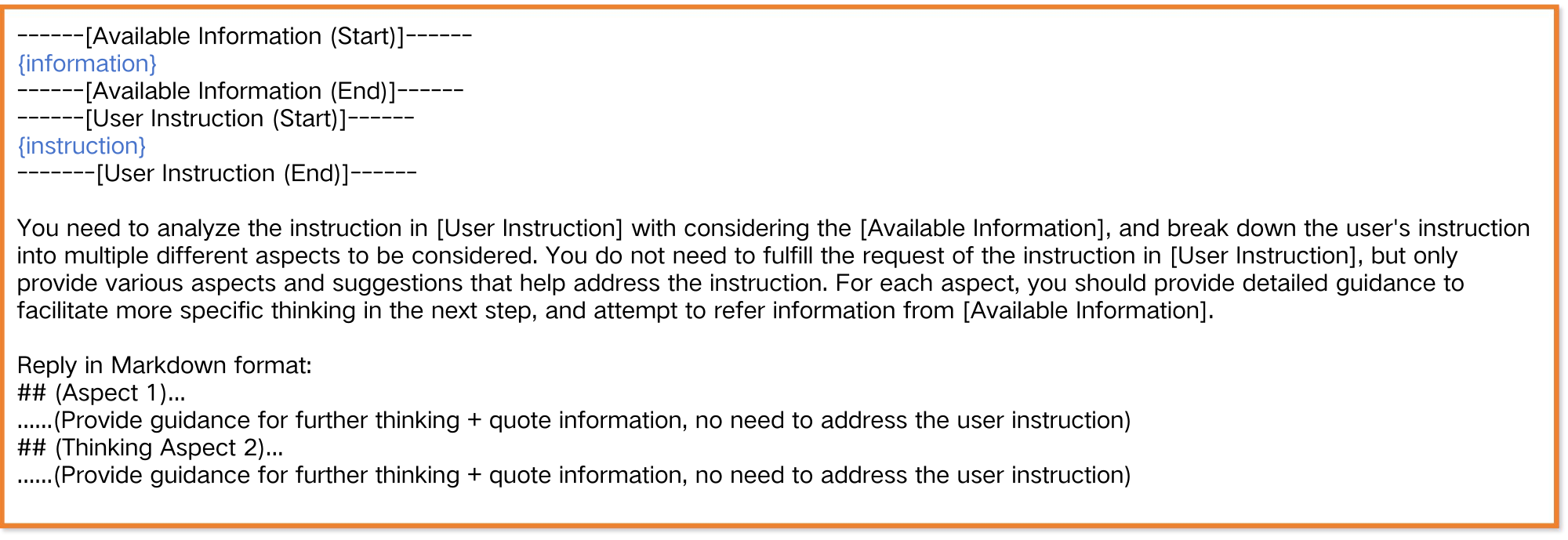}

\paragraph{Routing} If we apply these aspects for subsequent analysis following the naive wide-horizon paradigm, the planner's performance will exhibit severe scalability limitations as the number of aspects increases. Therefore, the strategist reorganizes these aspects and integrates them into a coherent planning blueprint. Here is the prompt used in the RFT and inference:

\includegraphics[width=1.0\linewidth]{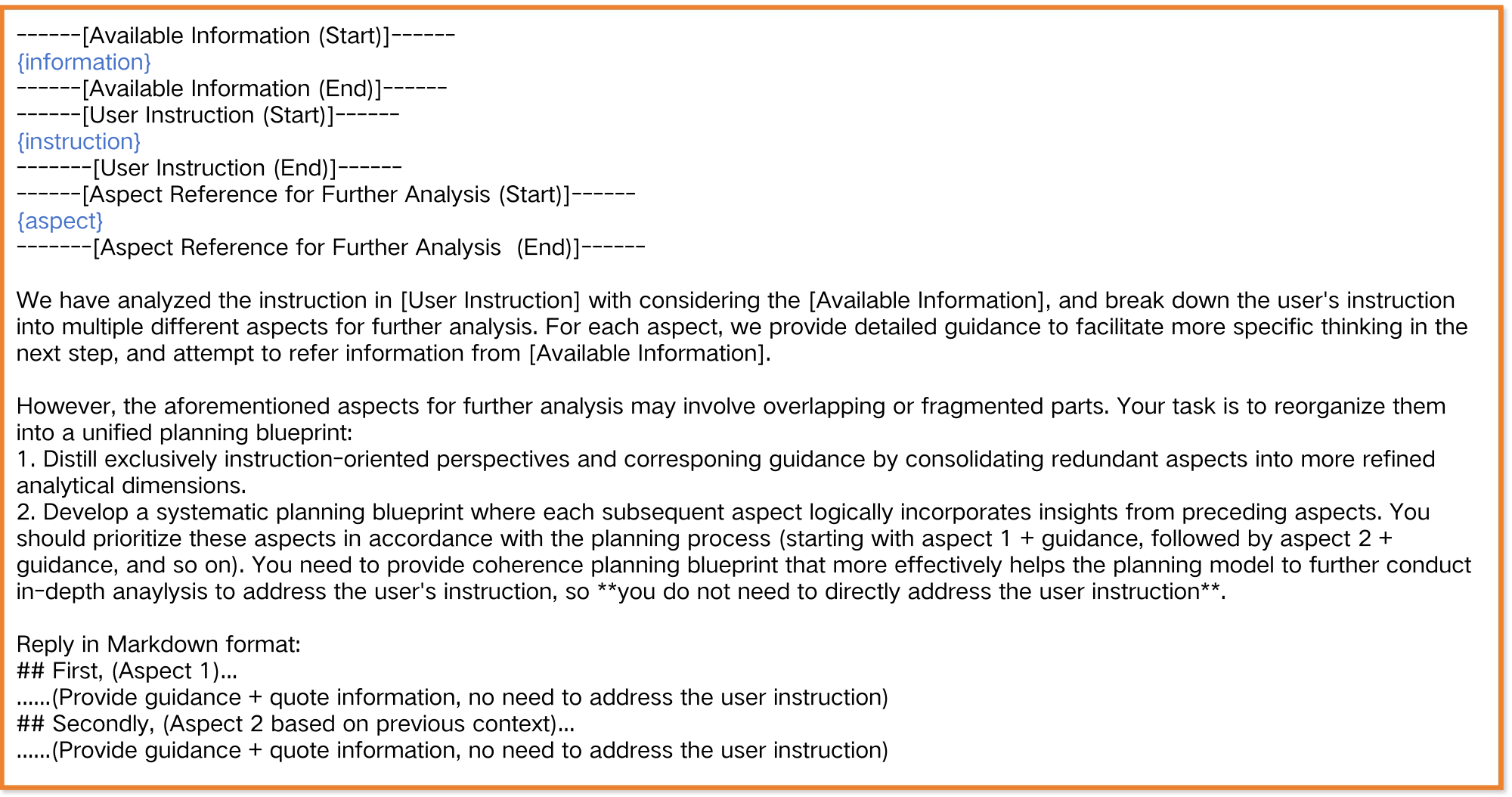}

\subsubsection{Inference for Planner}
\paragraph{Aspect-Aware Thinking} Based on the planning blueprint, the planner conducts a more focused and in-depth analysis following the order in the blueprint. The whole process of aspect-aware thinking is implemented through multiple turns of dialogue with shared history. We instruct the planner to follow the guidance to analyze one of the aspects in a turn of dialogue, as shown below:

\includegraphics[width=1.0\linewidth]{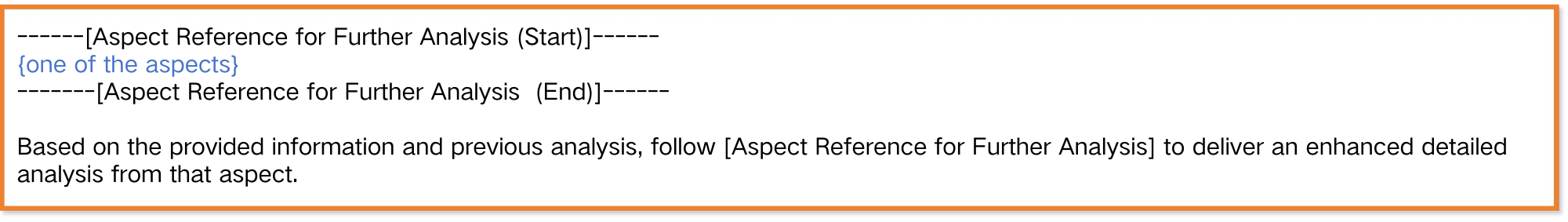}

\paragraph{Formatting Output Plan}
To integrate the previous analysis, we design formatting instructions for the model to output in a specified format. You can refer to the case study in Section \ref{app:case_simulation} to see an example of a plan that meets the formatting requirements.

\includegraphics[width=1.0\linewidth]{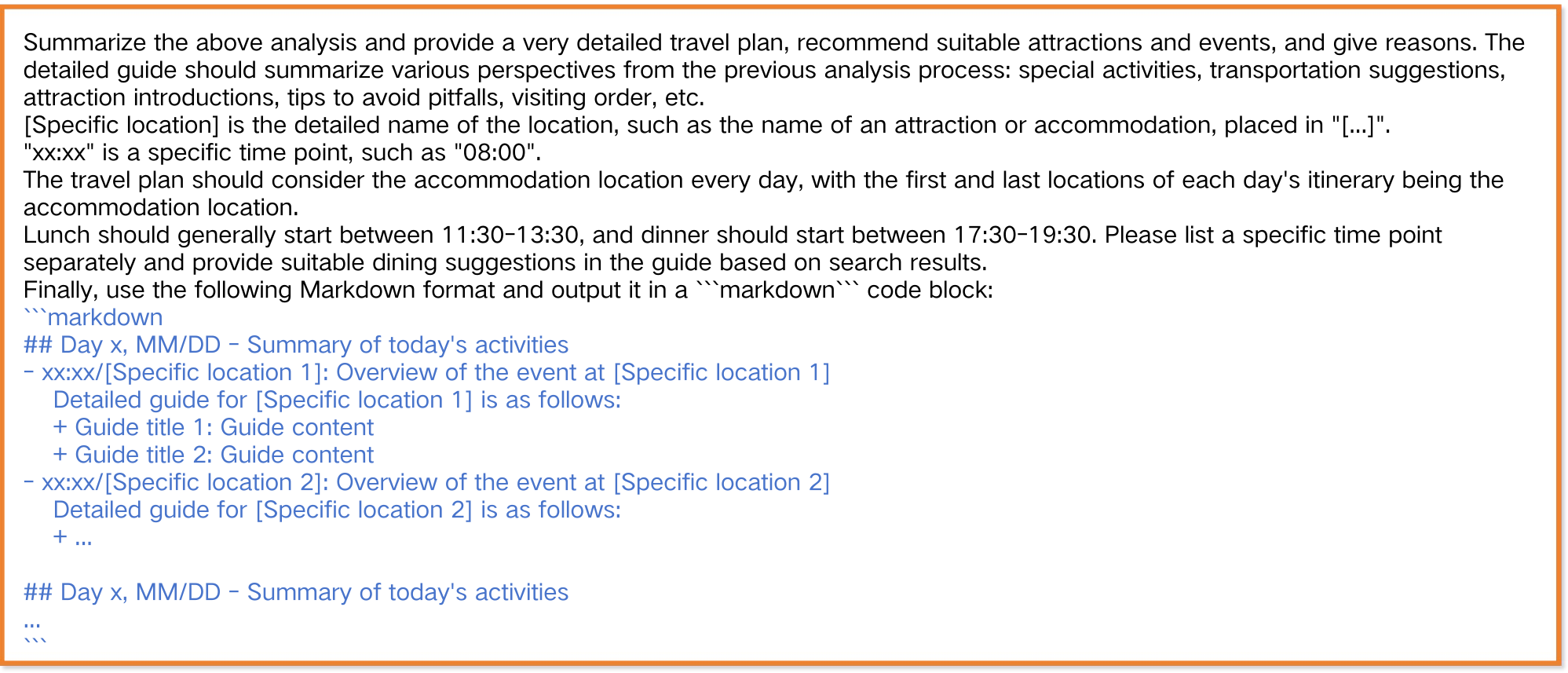}

\section{Simulation and Evaluation}
\label{app:travel_sim}
\subsection{Simulation Process}
\subsubsection{Action Space $\mathcal{A}$}
As we have mentioned, there are four kinds of actions permitted to transit to the next state: \emph{transiting}, \emph{resting}, \emph{dining}, and \emph{sightseeing}. In this section, we illustrate the details of how the travel agent takes these actions. 

After the traveler completes the event of the previous state, we prompt the traveler to make the decision for the next action as follows:

\includegraphics[width=1.0\linewidth]{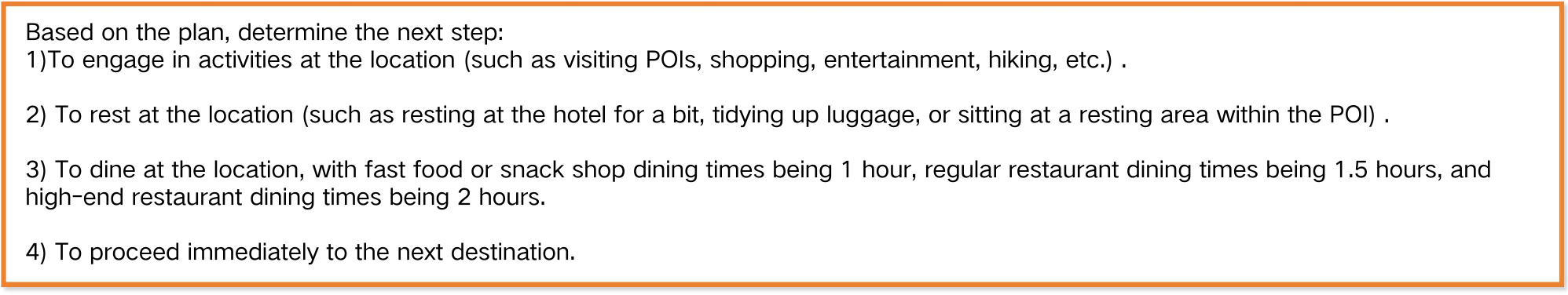}

Once the traveler chooses an action, we require the traveler agent to give the output formatted as follows:

\textit{
\textbf{Transiting: }\{
    "decision": "transit",
    "departure": "...",
    "destination": "...",
    "transport mode": "...",
    "arrival time": "xx:xx",
    "remaining stamina": ...,
    "total expense": ...,
    "next planned location": "..."
\}
}

\textit{
\textbf{Resting: }\{
    "decision": "rest",
    "end time": "xx:xx",
    "remaining stamina": ...,
    "total expense": ...,
    "next planned location": "..."
\}
}

\textit{
\textbf{Dining: }\{
    "decision": "dine",
    "end time": "xx:xx",
    "remaining stamina": ...,
    "total expense": ...,
    "next planned location": "..."
\}
}

\textit{
\textbf{Sightseeing: }\{
    "decision": "sightsee",
    "end time": "xx:xx",
    "remaining stamina": ...,
    "total expense": ...,
    "next planned location": "..."
\}
}

It is noted that only resting, which means resting in place, usually (If you're heading back to the hotel to rest, choose "transit"), does not need additional processing. The following actions, including transiting, dining, and sightseeing, need to be integrated with real-world information.

\paragraph{Transiting}
If the traveler chooses to transit to the next location according to the plan, there are several modes of transportation, including walking, cycling, public transportation (a combination of walking and bus/metro), and taxis. We use the Amap (Gaode) map API to obtain the travel time and costs of these modes of transportation. The traveler agent needs to comprehensively take into account time, cost, preferences, and stamina to make a choice. 

\paragraph{Dining}
If the traveler chooses to have a meal, we use the Amap (Gaode) map API to find the restaurant near the current location. We provide information on the quality and cost of restaurants for travelers to choose from, which is relevant to the traveler's preferences and budget. Dining usually takes 0.5-2 hours based on the restaurant type, and is also counted in the time of the sandbox environment.

\paragraph{Sightseeing}
If the traveler chooses to go sightseeing at the current location (usually a POI), we simulate the sightseeing experience of the POI. To be specific, we use the Red Note (Xiaohongshu) API to search for travel blog posts related to POI. Since the blog posts are based on real experiences, we design a sightseeing event agent to simulate the probable experience in advance, taking into account the traveler's preferences, available time, and stamina as follows:

\includegraphics[width=1.0\linewidth]{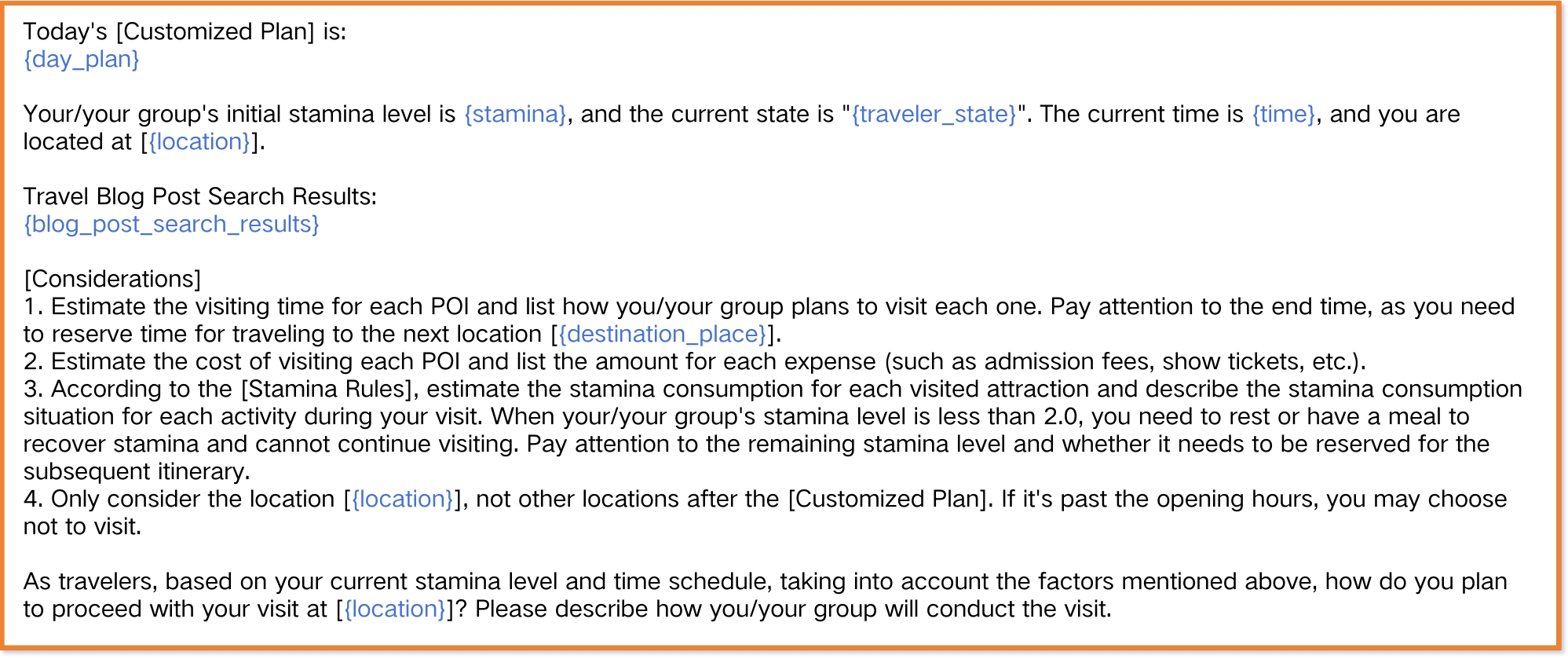}

Although we have simulated an experience of the POI in advance, the traveler can consider whether to actually conduct such an experience, considering the current time, stamina, or other factors.

\subsubsection{Traveler Stamina Engine}
\paragraph{Basic Design} Many researchers focus on analyzing the preference but neglect the type of traveler. The stamina of different types of travelers is diverse, which significantly influences the number of POIs to be visited in one day. In the travel simulation, we design the rule-based engine to adjust stamina expenditure and recovery, which especially vary for different types of travelers. Besides, each type of traveler possesses an initial stamina value, and the travel group is calculated as a whole.

As shown in Table \ref{tab:stamina}, we showcase some examples from Travel-Sim. For younger people, they have higher initial stamina, consume less energy, and recover faster. For elderly people, they would like to have a more relaxed journey and tend to be exhausted more quickly when sightseeing and transiting.

To enable LLMs to intuitively comprehend what stamina represents, we design a stamina-to-state conversion table that translates stamina values into specific states of the traveler: 1) If stamina is greater than 6.0, the state is "Energetic". 2) If the stamina is greater than or equal to 4.0 and less than 6.0, the state is "Good". 3) If the stamina is greater than or equal to 2.0 and less than 4.0, the state is "Slightly Tired". 4) If the stamina is less than 2.0, the state is "Very Tired".

\paragraph{Spontaneous Behavior} Although we do not explicitly instruct the LLM to be aware of the state of the traveler, we find it interesting that the traveler agent can \textbf{automatically adjust the next action based on the current state}. For example, if the traveler feels tired, the next action can be "resting in the restaurant" or "transiting by taxi instead of by bus to the next place". In some cases, the traveler modifies the itinerary or skips the next POI depending on the current state, which illustrates that stamina significantly influences the execution of the actual itinerary. This highlights that previous studies, failing to account for the varying stamina of travelers in planning, have resulted in travel plans that are less feasible.

\begin{table}[H]
\centering
\caption{Stamina rules for different types of travelers (from examples in Travel-Sim).}
\label{tab:stamina}
\fontsize{9}{11}\selectfont
\begin{tabular}{p{1cm}p{1.8cm}p{6cm}p{3cm}}
\toprule
\textbf{Type} & \textbf{Composition \{gender, age\}} & \textbf{[Initial stamina] Description} & \textbf{Stamina Rule} \\
\midrule
Single & \{male, 32\} & [8.5] An energetic 32-year-old male traveler loves to explore the natural scenery and historical culture, enjoys in-depth experience of local life through public transportation and walking, especially in traditional food and shopping for special souvenirs, preferring to avoid commercial attractions.
& sightseeing/1hr: -1; dining: +1; resting/1hr: +1; transiting/1h: bus/metro+0, taxi+0.5, walking-1, cycling-0.5, no cycling
\\
\midrule
Couple & \{male, 65\}, \{female, 62\} & [6.5] An elderly couple with a passion for culture and history, preferring a leisurely travel pace. They enjoy visiting temples and savoring local cuisine, all while prioritizing rest and comfort during their journeys. 
&  sightseeing/1hr: -1.5; dining: +0.5; resting/1hr: +0.5; transiting/1h: bus/metro-1, taxi+0, walking-1.5
\\
\midrule
Family & \{male, 32\}, \{female, 31\}, \{male, 7\}, \{female, 4\} & [7.0] A family of four consisting of a father who loves natural scenery and technology, a mother who loves shopping and searching for local cuisine, and two active children, they prefer a comfortable and convenient way to travel. The family focuses on parent-child experiences and cultural exploration.
& sightseeing/1hr: -1; dining: +0.5; resting/1hr: +0.5; transiting/1h: bus/metro-1, taxi+0, walking-1.5, no cycling
\\
\midrule
Family & \{male, 45\}, \{female, 41\}, \{female, 71\} & [6.5] A family of 45-year-old men, 41-year-old women, and 71-year-old women who prefer relaxing cultural and historical exploration, enjoying local cuisine and special performances, while focusing on the comfort of travel and quality of family time.
& sightseeing/1hr: -1.5; dining: +0.5; resting/1hr: +0.5; transiting/1h: bus/metro-1, taxi+0, walking-1.5, no cycling
\\
\midrule
Group & \{female, 23\}, \{female, 24\}, \{female, 27\} & [7.5] A group of three young female travelers who love to explore local food and culture, fashion shopping and photo check-in travel, prefer a relaxed and comfortable way, enjoy lively nightlife, and do not like to wake up early.
& sightseeing/1hr: -1; dining: +0.5; resting/1hr: +1; transiting/1h: bus/metro-0.5, taxi+0.5, walking-1, cycling-1 \\
\bottomrule
\end{tabular}
\end{table}

\subsection{Case Study}
\label{app:case_simulation}
In this section, we showcase an example to intuitively illustrate how the travel simulation works. First of all, we create a traveler agent based on an example in Travel-Sim, e.g., the elderly couple in Table \ref{tab:stamina}. We load the basic information, including the stamina rule and preferences, of the traveler to the system prompt of the LLM agent. 

This elderly couple is going to have a 3-day journey in Beijing. They start their journey by arriving at the Beijing West Railway Station at 10:00 AM. Their initial stamina will be subtracted by 2 ($6.5 - 2=4.5$) because of the exhaustion of traveling to a new city. The couple can choose one of the four actions to go to the next state. We offer four different routes utilizing various modes of transportation, all provided via the map API:

\includegraphics[width=1.0\linewidth]{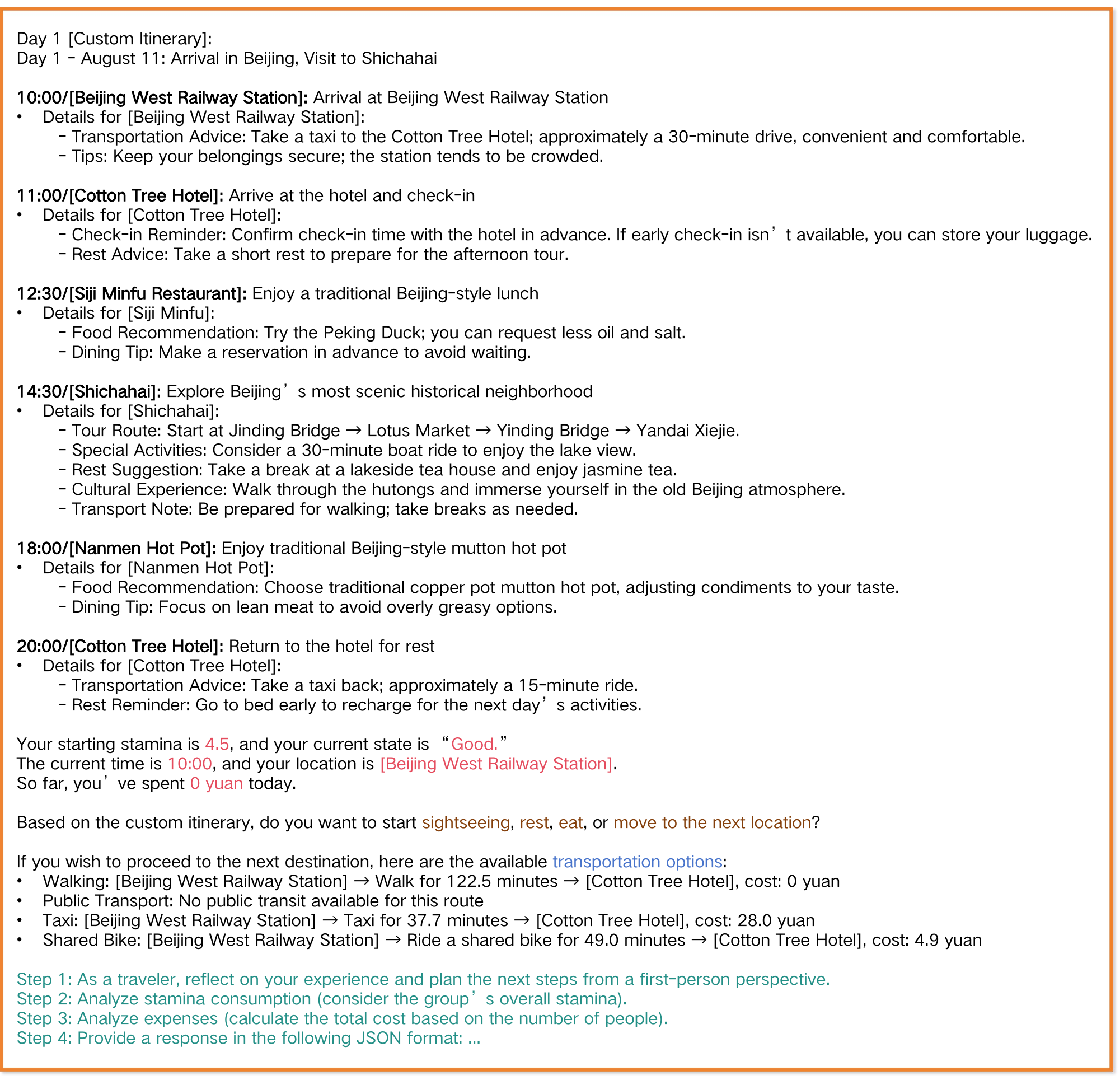}

Considering the travelers' perspective, they decide to take a taxi to their hotel for their next step, as they feel exhausted from the long journey to Beijing:

\includegraphics[width=1.0\linewidth]{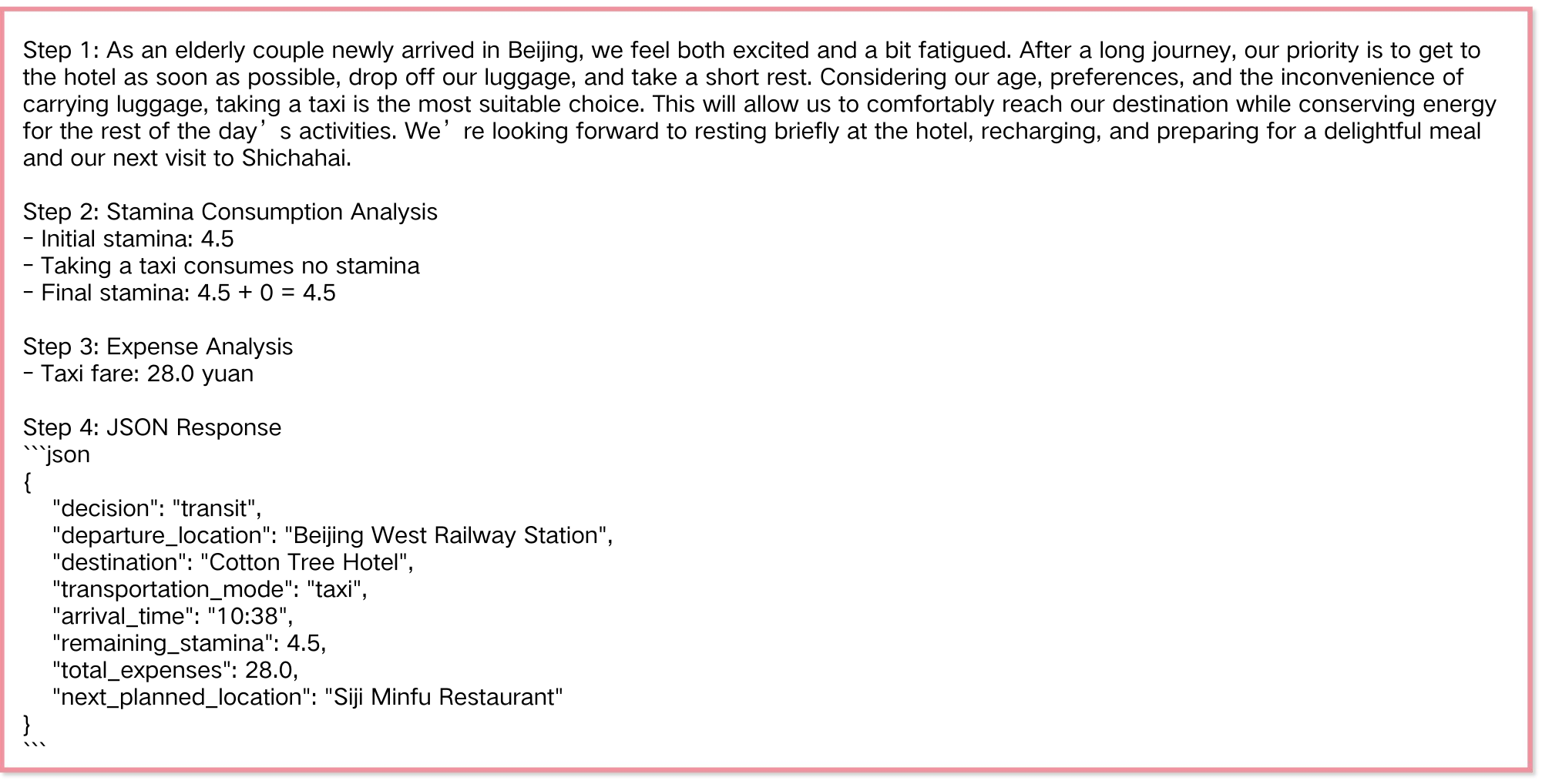}

After the couple arrives at the hotel, they choose to rest for a while before heading to the restaurant to taste the local cuisine, e.g., Peking duck:

\includegraphics[width=1.0\linewidth]{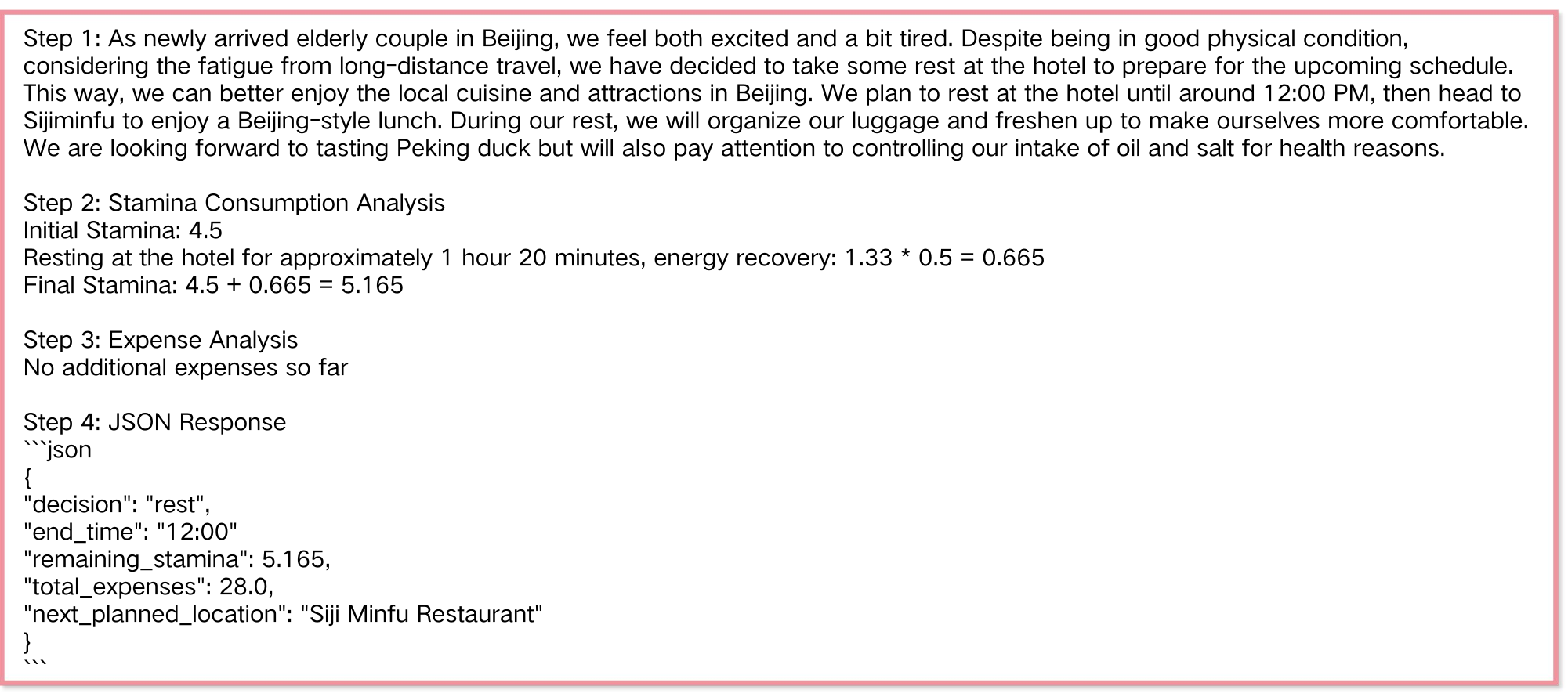}

(Skipping the states of transiting and dining)
After having a wonderful meal in the restaurant, they plan to visit their first attraction, Shichahai, a historic and scenic area in the heart of Beijing, encompassing three interconnected lakes: Qianhai (Front Sea), Houhai (Rear Sea), and Xihai (West Sea). Known for its well-preserved traditional Siheyuan courtyard houses and Hutongs, Shichahai offers visitors an authentic glimpse into old Beijing culture. To create a simulated sightseeing experience in Shichahai, we utilize the Red Note (Xiaohongshu) API to gather blog posts relevant to visiting this area. Based on the gathered content, we then guide the sightseeing event agent to craft an immersive experience from the perspective of the travelers:

\includegraphics[width=1.0\linewidth]{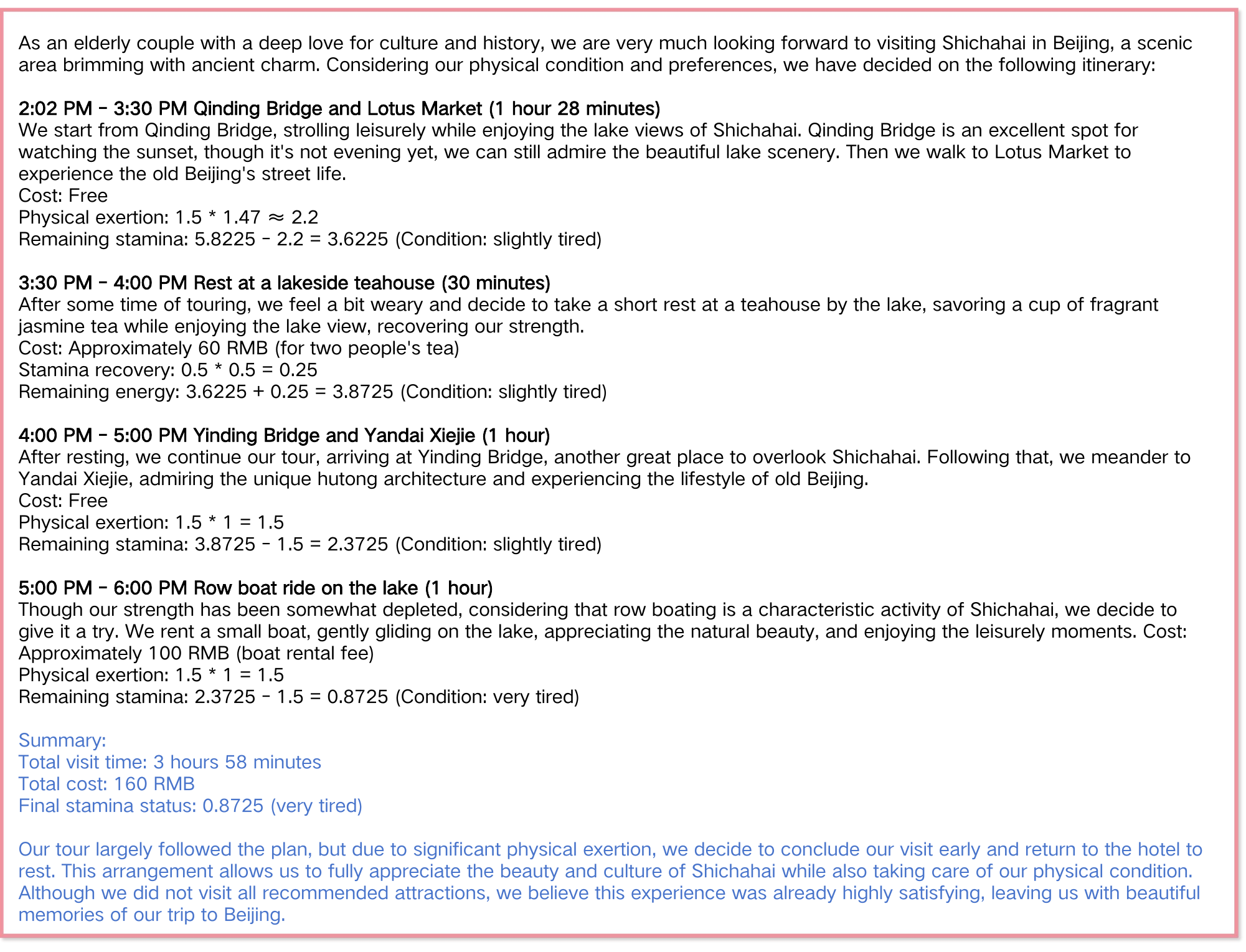}

Due to the limited space here, we present only a part of the simulation, but it encompasses most scenarios. This case study showcases that causal travel simulation is based on the integration of real-world information and comprehensively reflects the quality of the generated plan.

\newpage
\begin{figure}[ht]
    \centering
    \includegraphics[width=1.0\linewidth]{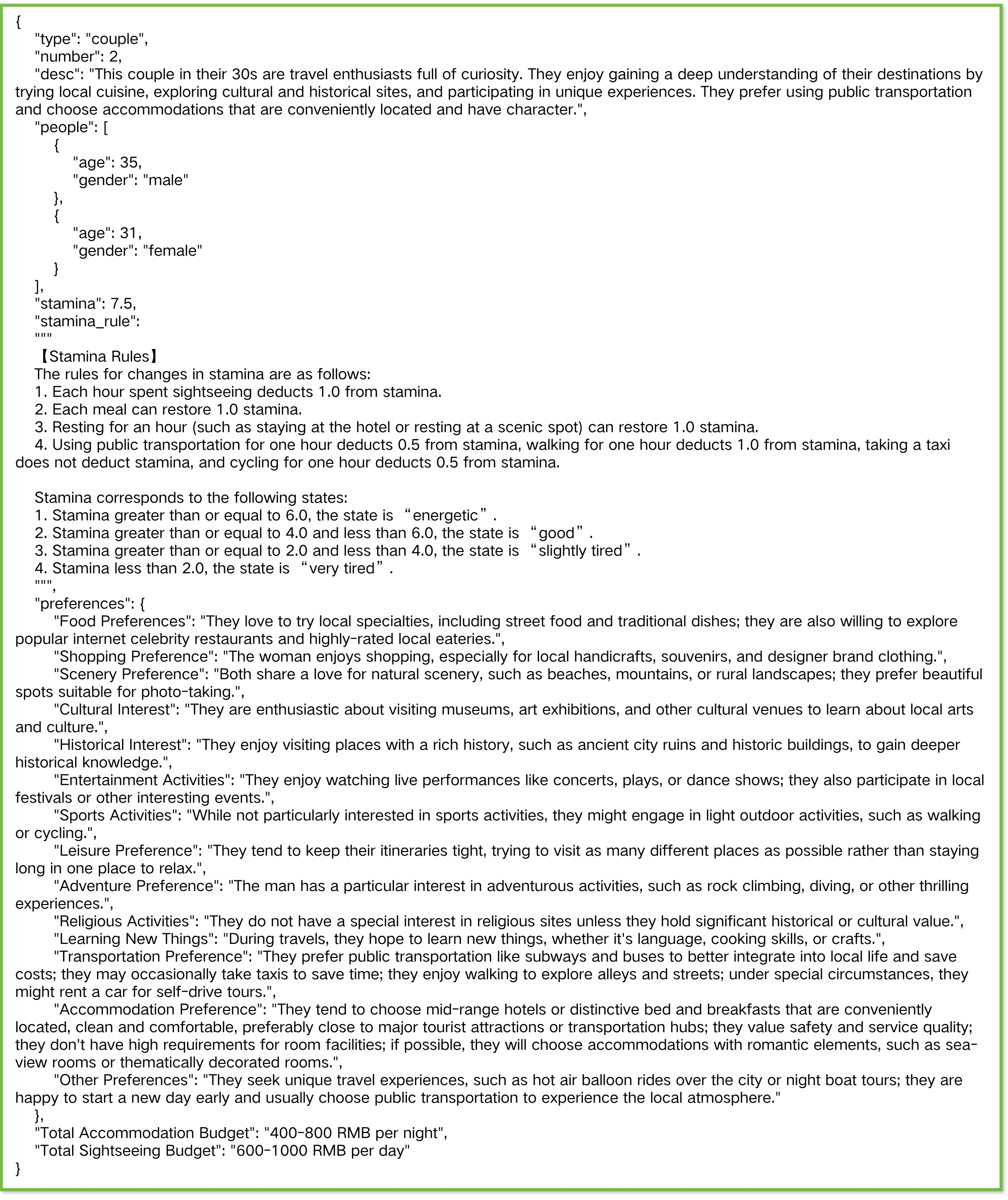}
    \caption{An example in Travel-Sim. The traveler group consists of a couple with one female and one male in their 30s. They have an initial stamina of 7.5 and consume less stamina for walking and cycling compared to elders and families with kids. They have specific and detailed preferences that will greatly influence their decision on the itinerary.}
    \label{fig:traveldata}
\end{figure}

\subsection{Travel-Sim Dataset Card}
\label{app:data}
As shown in the example in Figure \ref{fig:traveldata}, we construct the Travel-Sim dataset that consists of diverse travelers with detailed preferences. Different from previous travel planning benchmarks that only consider single travelers with limited preferences, Travel-Sim accommodates various types of travelers with diverse group compositions, such as individuals, couples, groups, families, and more. Each traveler type may have distinct preferences regarding activities, accommodation standards, budget constraints, travel pace, and cultural experiences. 

To build such a dataset, we annotate the profiles of 16 types of travelers and leverage Deepseek-R1 to expand the details. We select the 7 most-visited cities in China as the destinations and combine them with the travelers to generate 112 \{traveler, destination\} pairs as the evaluation dataset. 

\begin{table}[h!]
\centering
\caption{Datasets with City Composition.}
\label{tab:city}
\resizebox{\textwidth}{!}{%
\begin{tabular}{@{}llcc@{}}
\toprule
\textbf{Dataset} & \textbf{City List} & \textbf{City Count} & \textbf{Traveler Type Count} \\
\midrule
\textbf{Reward Model Training} & \begin{tabular}[t]{@{}l@{}}Guilin, Qingdao, Luoyang, Xishuangbanna, Shenyang, \\ Wuhan, Nanchang, Zhengzhou, Changchun, Xianyang, \\ Lanzhou, Yangzhou, Chaozhou, Guangzhou\end{tabular} & 14 & 37 \\
\midrule
\textbf{RFT/RL Training} & \begin{tabular}[t]{@{}l@{}}Beijing, Chongqing, Nanjing, Chengdu, Datong, \\ Jingdezhen, Dalian, Hangzhou, Beihai, Lijiang, \\ Kunming, Taiyuan, Xianning, Shenzhen, Hong Kong, \\ Changsha, Shantou, Qinhuangdao, Enshi, Tianjin\end{tabular} & 20 & 120 \\
\midrule
\textbf{Evaluation} & \begin{tabular}[t]{@{}l@{}}Lhasa, Sanya, Shanghai, Harbin, Dali, Xi'an, Xiamen\end{tabular} & 7 & 16 \\
\bottomrule
\end{tabular}%
}
\end{table}

\subsection{Multi-granularity Evaluation by Traveler}
\label{app:multig_eval}
After experiencing the itinerary, the traveler has much to share regarding it. We implement a multi-granularity evaluation mechanism for the travel experience from three levels: the traveler reflects on the experience and rates the score after each POI visit, at the day's conclusion, and after completing the entire journey. We first ask the traveler agent to think from the perspective of a traveler by outputting the psychological activities. We then ask the traveler agent to rate the score by inspecting five indicators, e.g., travel experience (ex), scenic spot characteristics (it), sightseeing arrangements (ar), stamina exertion (st), and overall expense (co).

\textbf{For every end of the visit to the POI}, we let the traveler assess the POI visiting experience as follows:

\includegraphics[width=1.0\linewidth]{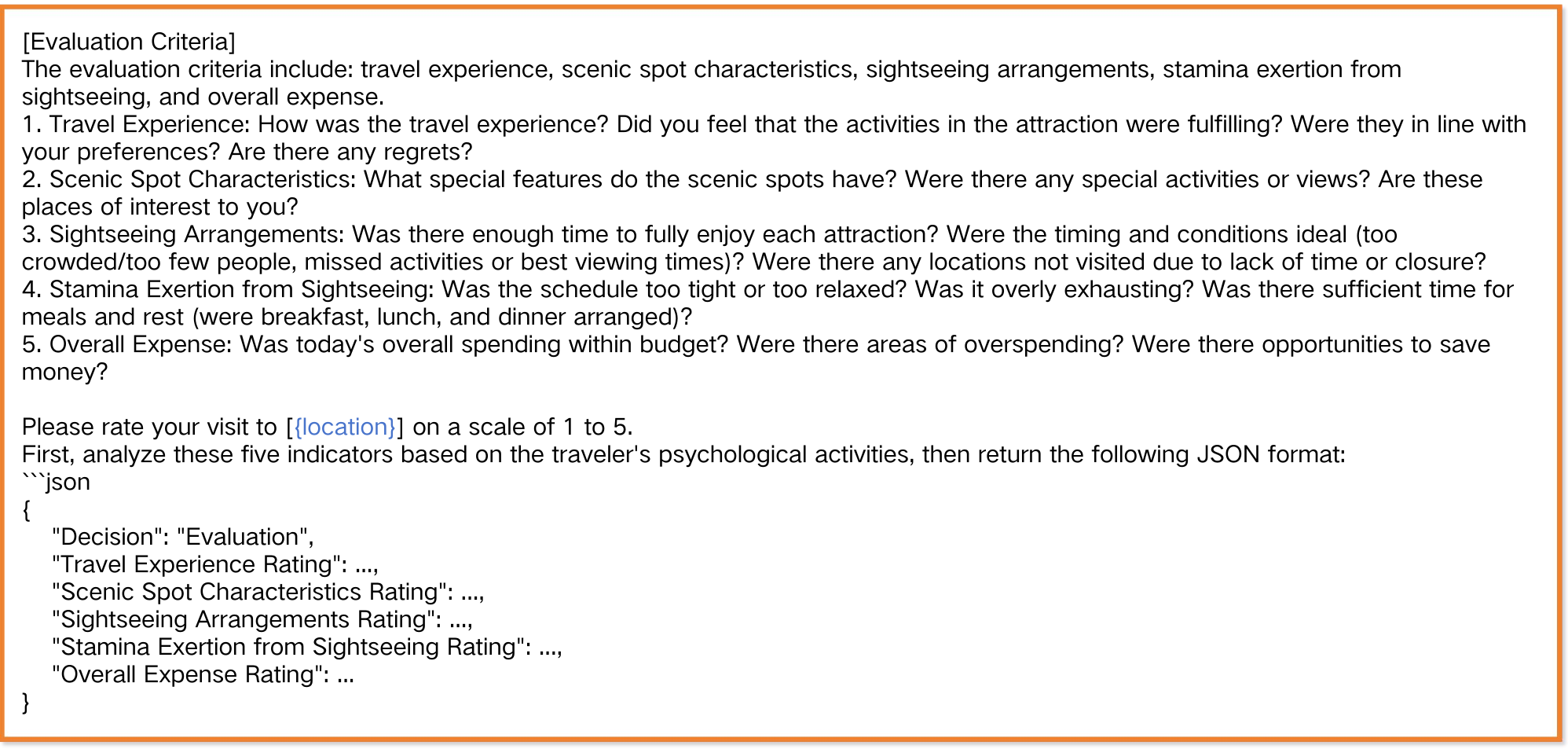}

\textbf{For every end of the whole-day itinerary}, we let the traveler assess the travel experience of today as follows:

\includegraphics[width=1.0\linewidth]{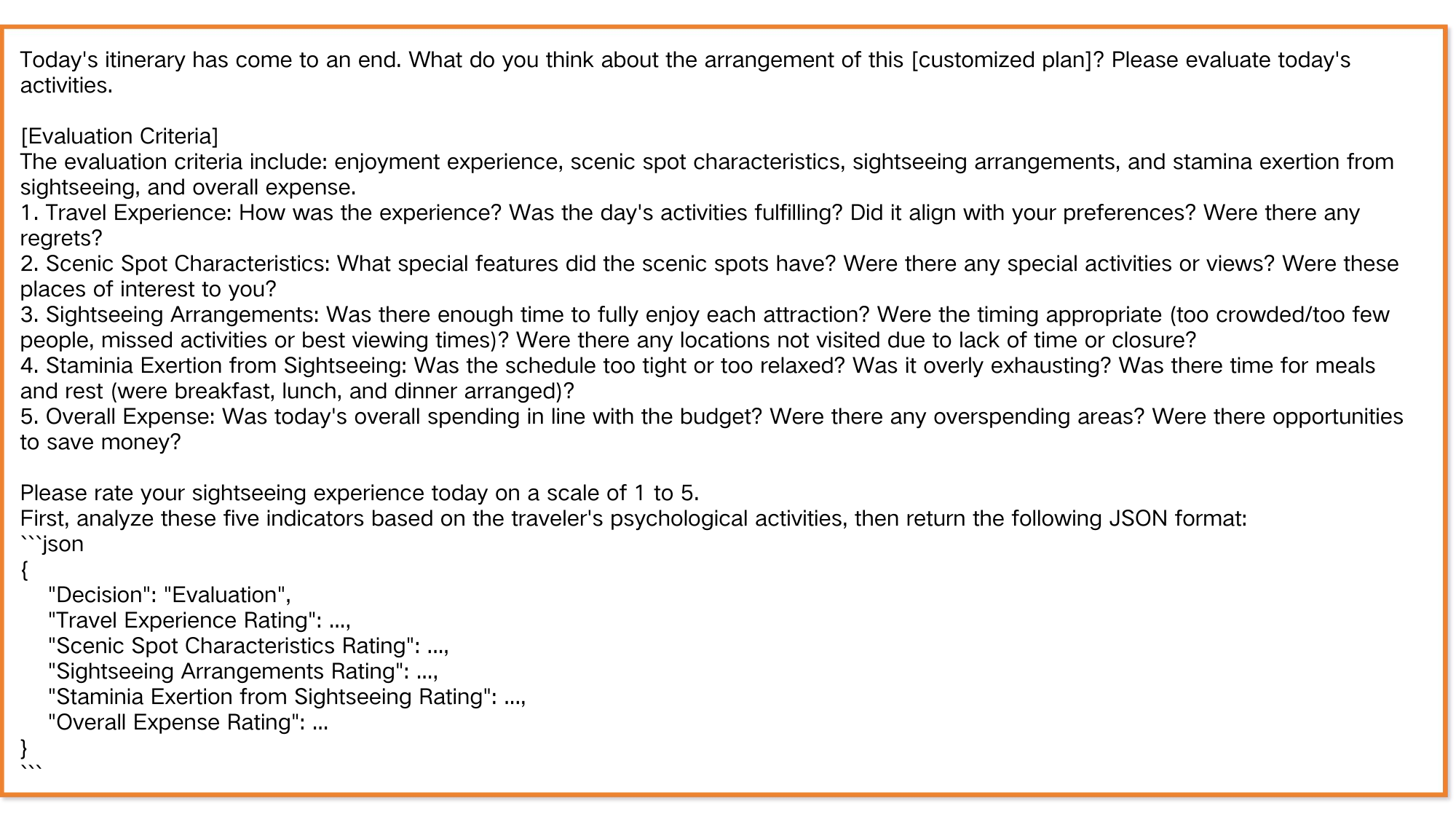}

\textbf{At the end of the multi-day journey}, we let the traveler assess the overall travel experience of the entire trip as follows:

\includegraphics[width=1.0\linewidth]{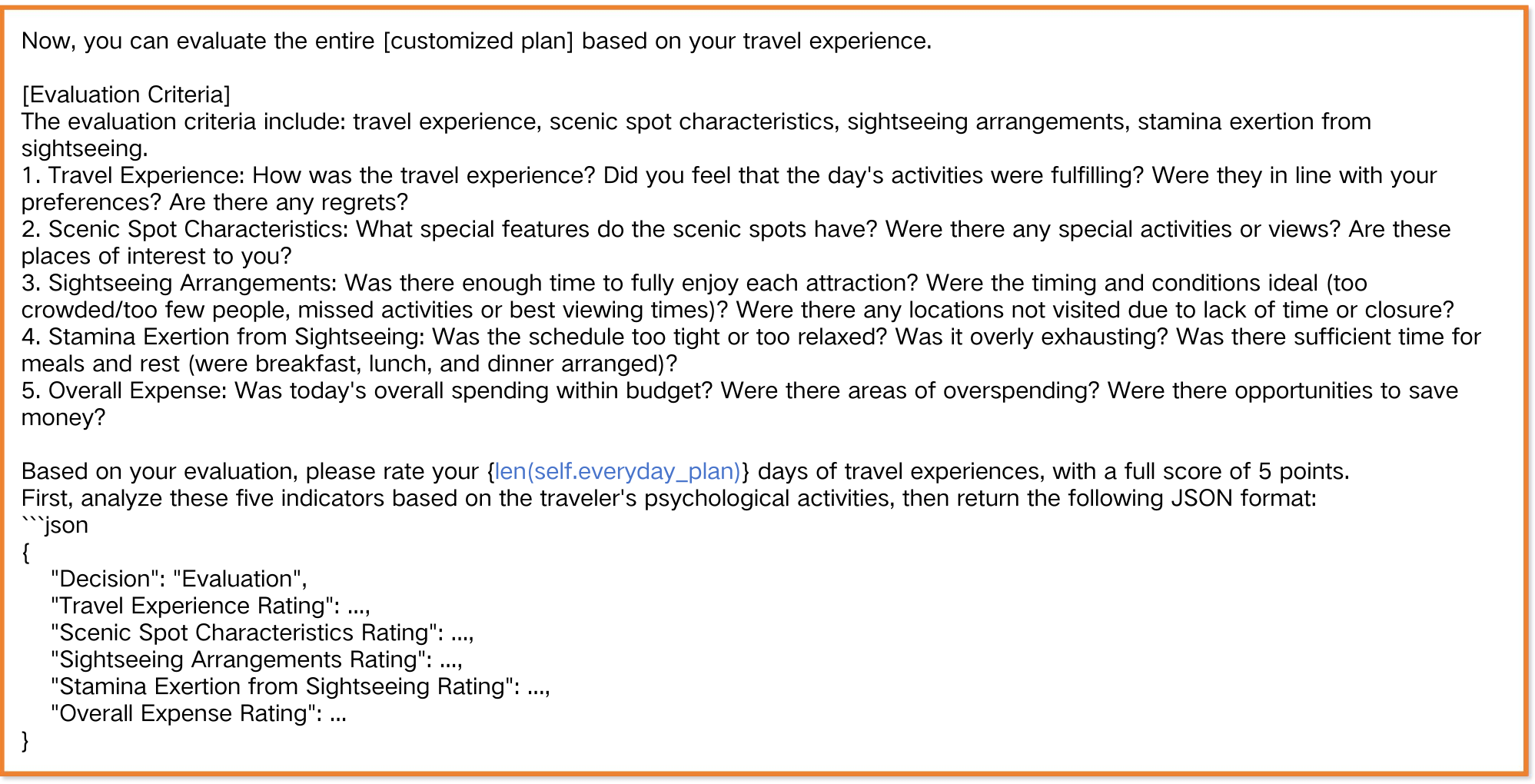}


\subsection{Human Verification for Simulation-based Evaluation}
\label{app:human}
Although the simulated travel empirically seems to be effective for evaluating the generated plan, we further implement human evaluation to verify if the simulation-based evaluation is consistent with the human evaluation. To be specific, we provide the evaluation results from the simulated travel for humans to check if the evaluation from the traveler agent is reasonable. For example, given the evaluation of the POI travel experience, humans inspect it based on the map and blog post, subsequently deciding whether to endorse the evaluation. If humans agree with the travel agent's evaluation, the scores remain the same as those of the travel agent. If humans disagree, they have the option to modify the scores. 

Three individuals take part in this experiment. We record the modified scores and the deviation values to calculate the agreement rates. It is noted that the scores here are PER-ex, PER-it, PER-ar, PER-st, and PER-co. 

As shown in Table \ref{tab:human}, the evaluation of traveler agents has high consistency with the human evaluation, with the agreement rate of 92\%. We have identified that the PER-co (travel cost) metric exhibits a relatively higher level of inconsistency. The travel agent often buys too many expensive souvenirs, which leads to going over budget. Additionally, they occasionally make mistakes when calculating total travel expenses, especially when accounting for a group of travelers.

\begin{table}[H]
\centering
\caption{We implement human evaluation to verify if the simulation-based evaluation is consistent with humans'.}
\label{tab:human}
\setlength{\tabcolsep}{5pt} 
\renewcommand{\arraystretch}{1.3} 
\fontsize{9}{11}\selectfont
\begin{tabular}{lccccccc}
\toprule
& ex & it & ar & st & co & PER-agg. & Agree. rate\\
\midrule
R1-Distill 7B (s.) + R1-Distill 7B (p.) 
& 77.5 & 86.4 & 79.3 & 76.4 & 87.6 & 81.4 & - \\
Human 
& 69.2 & 83.7 & 73.4 & 74.2 & 74.5  & 75.0 & 92.1\% \\
\bottomrule
\end{tabular}
\end{table}

\section{Metrics}
\label{app:metric}
We examine four criteria to evaluate the effectiveness of wide-horizon thinking. Comprehensiveness (CPH) and Completeness (CPL) are rule-based metrics, while Feasibility (FEA) and Personalization (PER) are simulation-based metrics.

\subsection{Rule-based Metrics}
\subsubsection{Comprehensiveness (CPH)}
Comprehensiveness (CPH) evaluates how much relevant information is effectively integrated from the long context into the final plan. To elaborate, for each POI in the plan, we first extract the corresponding travel guidance. We calculate the similarity between the POI travel guidance and POI-related blog posts in the context. We encode both texts into embedding and calculate the cosine similarity \footnote{We use the \emph{paraphrase-multilingual-mpnet-base-v2} model of the \emph{sentence-transformers} library in \url{https://sbert.net/docs/sentence_transformer/pretrained_models.html}.}. We calculate the average similarity across all POIs in the plan as the comprehensive score.

\subsubsection{Completeness (CPL)}
To create an organized travel plan, we must include several essential elements, such as the timeline, destinations, activities, etc. We require LLMs to generate a travel plan that follows the specific output structure and includes the required elements. To evaluate if the generated plan strictly follows formatting instructions for a thorough itinerary, we inspect four criteria as follows:
\begin{enumerate}
    \item The origin and destination of the entire journey must be at the specified stations or airports.
    \item Each day's itinerary, excluding the first and last day, will begin and end at the hotel where the traveler is accommodated.
    \item The introduction for each point of interest's sightseeing should be formally structured according to the format specified in the prompt.
    \item In the travel itinerary, activities should be scheduled to include meal arrangements for both lunch and dinner.
\end{enumerate}
For criteria 1, 2, and 3, we use the regex expression to extract the keywords and verify whether the generated plan satisfies these criteria. For the last one, we leverage the Deepseek-R1 \cite{guo2025deepseekr1} to verify if the itinerary of each day includes arrangements for lunch and dinner. We evaluate each plan against these four criteria, awarding 25 points for each criterion that is met. Therefore, for each plan, we have a maximum of 100.0 points for CPL by summing all the scores.

\subsection{Simulation-based Metrics}
Because some criteria, e.g., feasibility and personalization, are hard and less persuasive to be simply measured by rule-based metrics, we deal with them via travel simulation based on the real world.
\subsubsection{Feasibility (FEA): Travel Plan Similarity Score}
We evaluate the feasibility of the generated plan by ensuring that the plan is realistic and executable within the given constraints, such as time and spatial optimization. As shown in Figure \ref{fig:trajectory}, we can perceive the generated travel plan as a trajectory of multiple "time-location" pairs. As the traveler conducts a virtual journey based on the generated travel plan, the traveler also produces a trajectory of "time-location" pairs. We aim to assess the feasibility of the generated plan by evaluating whether the plan aligns with the traveler's itinerary. 

To calculate the similarity between two trajectories of "time-location" pairs, there are a few things to consider: 
\begin{enumerate}
    \item \textbf{Completeness}: The events of two trajectories should match. Missing or extra events should be penalized.
    \item \textbf{Chronological order}: The events of two trajectories should happen in the same order.
    \item \textbf{Temporal proximity}: The identical events of these two trajectories should occur as closely in time as possible. A penalty shall be imposed that corresponds to the temporal discrepancy between identical events. The magnitude of this penalty increases proportionally with the extent of the time gap.
\end{enumerate}


Based on the considerations above, as shown in Algorithm \ref{alg:tpss}, we design an algorithm for calculating the similarity between two trajectories of "time-location" pairs, dubbed \textbf{Travel Plan Similarity Score (TPSS)}. We separately deal with the similarity of time and locations and implement dynamic programming to iteratively calculate the score. It is noted that we use TPSS to calculate the similarity of trajectories of one day. For multi-day journeys, we calculate the final TPSS by averaging the daily scores.

\begin{algorithm}[H]
\caption{Travel Plan Similarity Score Calculation}
\label{alg:tpss}
\begin{algorithmic}
\State \textbf{Input:} generated\_plan\_trajectory $T_g$, simulated\_travel\_trajectory $T_s$
\State \textbf{Output:} similarity score (as percentage)

\Function{CalculatePlanSimilarity}{$T_g$, $T_s$}
    \State $m \gets$ length of $T_g$
    \State $n \gets$ length of $T_s$
    \State Initialize $dp[m+1][n+1]$ with zeros
    
    \For{$i \gets 1$ to $m$}
        \For{$j \gets 1$ to $n$}
            \State $score \gets$ CalculateMatchScore($T_g$[i-1], $T_s$[j-1])
            \State $dp[i][j] \gets \max(dp[i-1][j], dp[i][j-1], dp[i-1][j-1] + score)$
        \EndFor
    \EndFor
    
    \State $max\_score \gets \min(m, n)$ \Comment{Ideal maximum score}
    \State $similarity \gets dp[m][n] / max\_score$
    \State $completeness\_penalty \gets \min(m, n) / \max(m, n)$ \Comment{Penalize for missing/extra activities}
    \State $final\_similarity \gets similarity * completeness\_penalty * 100$
    
    \State \textbf{return} $final\_similarity$
\EndFunction

\Function{CalculateMatchScore}{item1, item2}
    \State $time\_score \gets$ TimeDiffScore(item1.time, item2.time)
    \If{item1.location = item2.location}
        \State $location\_score \gets 1$
    \Else
        \State $location\_score \gets 0$
    \EndIf
    \State \textbf{return} $(time\_score + location\_score) / 2$
\EndFunction

\Function{TimeDiffScore}{time1, time2}
    \State $diff \gets |$time2 - time1$|$ in hours
    \State \textbf{return} $\max(0, 1 - diff / 2)$ \Comment{Linear decrease within 2 hours}
\EndFunction
\end{algorithmic}
\end{algorithm}

\subsubsection{Personalization (PER): Aggregated Assessment from Travelers}
We evaluate the personalization of the generated plan by inspecting whether the travel plan meets the unique needs and preferences of the traveler agent in the simulation. As we have mentioned, our multi-granularity evaluation mechanism via traveler's assessment in Appendix \ref{app:multig_eval}, we further introduce how we aggregate the scores from multi-granularity assessment in Equation \ref{eq:per} as follows:
\begin{equation}
\label{eq:per}
    \mathcal{S}_{PER} = \alpha_1 \mathcal{S}_{whole\_travel} + \alpha_2\frac{1}{N}\sum_i^N (\beta\mathcal{S}_{day\_i} + \gamma \mathcal{S}_{POI\_day\_i}).
\end{equation}
As each POI has a score after sightseeing, we average the scores of each POI to be the overall POI score $\mathcal{S}_{POI\_day\_i}$ in the $i^{th}$ day. We weight the $\mathcal{S}_{POI\_day\_i}$ and the score of the whole-day itinerary $\mathcal{S}_{day\_i}$ by $\beta$ and $\gamma$  as the aggregated score of $i^{th}$ day. We set $\beta$ as 0.6 and $\gamma$ as 0.4. We average the aggregated score of each day and weight it with the score of the whole travel by $\alpha_1$ and $\alpha_2$. We set $\alpha_1$ as 0.6 and $\alpha_2$ as 0.4. We normalize the $\mathcal{S}_{whole\_travel}$ to 100.0 as the final score.

\end{document}